\documentclass[11pt, english]{article}

\usepackage[margin=25truemm]{geometry}
\usepackage{authblk}
\usepackage[sort&compress,numbers]{natbib}
\usepackage{url}            
\usepackage{bm}
\usepackage{textcomp}
\usepackage{booktabs}       
\usepackage{amsfonts}       
\usepackage{nicefrac}       
\usepackage{microtype}      
\usepackage{amsmath}
\usepackage{amsthm}
\usepackage{mathtools}
\usepackage{mathrsfs}
\usepackage{subcaption}
\usepackage{graphicx}
\usepackage{cancel}
\usepackage{caption}
\usepackage{color}
\usepackage[colorlinks=true, allcolors=blue]{hyperref}

\newcommand{\iprod}[1]{\left\langle#1\right\rangle}

\newcommand{\verbar}[1]{\left|#1\right|}

\newcommand{\parentheness}[1]{\left(#1\right)}

\newcommand{\curly}[1]{\left\{#1\right\}}

\newcommand{\RR}{\mathbb{R}}

\newcommand{\EE}{\mathbb{E}}
\newcommand{\TT}{\mathbb{T}}

\newcommand{\ZZ}{\mathbb{Z}}

\newcommand{\dd}{\mathrm{d}}

\newtheorem{thm}{Theorem}[section]
\newtheorem{prop}{Proposition}

\title{Deep Learning in Random Neural Fields: Numerical Experiments via Neural Tangent Kernel}

\author{Kaito Watanabe$^1$, Kotaro Sakamoto$^2$, Ryo Karakida$^3$, Sho Sonoda$^4$, Shun-ichi Amari$^{5,6,7}$}

\affil{$^1$LPIXEL Inc, Japan \\
       $^2$The Institute of Statistical Mathematics, Japan \\
       $^3$AIST, Japan \\
       $^4$RIKEN AIP, Japan \\
       $^5$RIKEN CBS, Japan \\
       $^6$ACRO, Teikyo University, Japan \\
       $^7$Araya Inc, Japan}

\begin{document}

\maketitle

\begin{abstract}
A biological neural network in the cortex forms a neural field. Neurons in the field have their own receptive fields, and connection weights between two neurons are random but highly correlated when they are in close proximity in receptive fields. In this paper, we investigate such neural fields in a multilayer architecture to investigate the supervised learning of the fields. We empirically compare the performances of our field model with those of randomly connected deep networks. The behavior of a randomly connected network is investigated on the basis of the key idea of the neural tangent kernel regime, a recent development in the machine learning theory of over-parameterized networks; for most randomly connected neural networks, it is shown that global minima always exist in their small neighborhoods. We numerically show that this claim also holds for our neural fields. In more detail, our model has two structures: i) each neuron in a field has a continuously distributed receptive field, and ii) the initial connection weights are random but not independent, having correlations when the positions of neurons are close in each layer. We show that such a multilayer neural field is more robust than conventional models when input patterns are deformed by noise disturbances. Moreover, its generalization ability can be slightly superior to that of conventional models.
\end{abstract}

\section{Introduction}
Neural networks in the brain cortex have been optimized to perform intelligent information processing throughout their long evolutionary history. The aim of theoretical neuroscience is to understand the complex dynamics of neural networks such as pattern formation \cite{amari1977dynamics}, self-organized feature extraction \cite{kohonen1982self}, and topographic organization \cite{takeuchi1979formation}. Complex dynamics of neural networks have been studied by using two types of models: random models and field models. The random model is a neural network model with random connections. Theoretical analyses of this model are known as statistical neurodynamics or mean field theory, and they have enabled us to understand the complicated dynamics of neural networks through macroscopic parameters \cite{rozonoer1969random, amari1971characteristics, amari1972characteristics, amari1974method, amari1977mathematical, sompolinsky1988chaos, poole2016exponential, schoenholz2017deep, karakida2019universal, karakida2021pathological}. Random models are also used in practical applications, for example, the echo state network in reservoir computing \cite{jaeger2004harnessing}. On the other hand, in a neural field, neurons form a continuous field, capturing the structure of the cortex, which consists of layered sheets of densely aligned neurons \cite{weiner1946mathematical, beurle1956properties, wilson1972excitatory, amari1972learning, wilson1973mathematical, amari1977dynamics, takeuchi1979formation, kohonen1982self, Faugeras2012neural, coombes2014tutorial, touboul2014propagation, sonoda2017double}. 

While these theoretical models have shed light into the firing dynamics of neural networks, it has been unclear how they can be combined with problems of learning. In particular, there have been few studies on the supervised learning of neural fields \cite{brady2019neural}. Recently, however, in the literature of deep learning, studies on the {\it neural tangent kernel (NTK)} \cite{jacot2018neural} have revealed that the random model can achieve zero training error and a high generalization performance with a sufficiently small change in parameters. NTK research has shown that the global optimum for a given set of training examples is always found in a small neighborhood of randomly assigned initial connections \cite{jacot2018neural, amari2020any}. Within the NTK regime, a number of studies have shed light on the global convergence and generalization properties of sufficiently wide neural networks \cite{xie2017diverse, chizat2018lazy, li2018learning, du2018gradient, du2018gradient2, lee2019wide, Arora2019OnEC, allen2019learning, arora2019fine, yang2019scaling, cao2019generalization, zou2019gradient}. From this, we can expect that if the neural field model can be merged with the random model, we can formulate the supervised learning of the neural field through the NTK framework. 

Inspired by these trends of research, we investigate the robustness and generalization ability of multilayer neural fields in supervised learning tasks. Our model of multilayer neural fields has a correlated initial random structure of neural connections in which each neuron has a continuous receptive field. A typical deep neural network model does not use correlations among initial connections. It is physiologically known that the activities of neighboring neurons are correlated \cite{cohen2011measuring}. Mathematically, this means reduction in the size of the reproducing kernel Hilbert space (RKHS). It is known that the generalization performance can be improved by reducing the RKHS \cite{bietti2019group}. Thus, we implement these structures with continuously distributed receptive fields by introducing weights depending on the distances between neurons and random initial connection weights that are not statistically independent but have correlations depending on the positions of the neurons in neural fields. In the case of no correlations and sufficiently large receptive fields, our model simply reduces to a conventional multilayer neural network. 

In the present paper, we first formulate our model mathematically. Next, we address three research questions: (1) How do the performances of neural fields depend on the intensity of the correlation in the initial random connections and also the size of the receptive field? (2) Are neural fields with correlated neurons and receptive fields still governed by the NTK regime? (3) Does our model improve robustness to perturbations? We aim to investigate these questions with numerical simulations, although the simulations are preliminary.

\vspace{0.8em}
\noindent
\textbf{Contributions.} The main results in the present paper are as follows. 
\begin{itemize}
\setlength{\leftskip}{-0.6cm}
    \item We formulate and investigate the supervised learning of multilayer random neural fields. We numerically confirm that our model of multilayer random neural fields with correlated neurons and receptive fields is governed by the NTK regime.
    \item We find that our model of random neural fields is robust under the disturbances of both random noise and deformation of training samples. The generalization ability is slightly superior to those of conventional models of neural networks.
\end{itemize}

\section{Random Neural Fields}
For simplicity, we assume periodic boundary conditions on a one-dimensional neural field. Namely, the field is formulated as a function on a one-dimensional torus $\mathbb{T} := \mathbb{R}/T\mathbb{Z}$ with some period $T$. Here, $\mathbb{T}$ is identical to the circle $\mathbb{S}^1$, and obtained from an interval $[-T/2,T/2]$ by gluing the opposite sides together. This setting is widely used in the study of neural fields \cite{takeuchi1979formation, bressloff2011spatiotemporal, rankin2014continuation} and improves the mathematical outlook. In this paper, we adopt a one-dimensional torus $\TT$ for the boundary condition, which is the line segment $[0,1]$ with both ends identical.

\subsection{Formulation}
We formulate a $d$-dimensional random neural field ($d$-RNF). Typically, $d=2$ for image processing. The positions of neurons in the $l$th layer are denoted by $\bm{z}_{l} \in \TT^d$. The behavior of the model is described as
\begin{align}
    \begin{cases}
        {h}^{l}(\bm{z}_{l}) &=  \displaystyle\int_{\TT^d} {w}^{l}(\bm{z}_{l},\bm{z}_{l-1}){x}^{l-1}(\bm{z}_{l-1})\dd\bm{z}_{l-1} + {b}^{l}(\bm{z}_{l}) \\
        {x}^{l}(\bm{z}_{l}) &= \varphi ( {h}^{l}\left( \bm{z}_{l} ) \right),
    \end{cases}
\end{align}
where ${w}^{l}(\bm{z}_{l},\bm{z}_{l-1})$ is a scalar weight connecting a neuron at position $\bm{z}_{l-1}$ in the $(l-1)$-th layer to a neuron at position $\bm{z}_{l}$ in the $l$th layer, $h^l(\bm{z}^l)$ is the pre-activation of the neuron at $\bm{z}_l$, ${b}^{l}(\bm{z}_{l})$ is a bias function, and ${x}^{l}(\bm{z}_{l})$ is the output of this neuron, which becomes the input to the next $(l+1)$-th layer. Here, $\varphi\left(\cdot\right)$ is an activation function.

The final output $y \in \mathbb{R}$ is a scalar that is a linear function of the outputs of the final layer $L$ of the neural field for input $\bm{x}=\bm{x}^0$,
\begin{align}
    y = f(\bm{x},\theta) =  \int_{\TT^d} {w}^{L+1}(\bm{z}_{L}){x}^{L}(\bm{z}_{L})\dd\bm{z}_{L}.
\end{align}
We denote by vector $\theta$ the set of all parameters: $\theta = \{ {\tilde{w}}^{l}(\cdot,\cdot), {b}^{l}(\cdot); \ l=1,\dots,L+1\}$.

The initial weights and biases are distributed according to a Gaussian process as $\mathcal{GP}(0,K)$ and $\mathcal{GP}(0,K_b)$ with mean zero and covariance function $K$ and $K_b$ depending on the positions of neurons in the $(l-1)$-th and $l$-th layers, as
\begin{align}
\label{wab}
  \begin{cases}
    \tilde{w}^l(\bm{z}_l,\bm{z}_{l-1}) \sim \mathcal{GP}(0,K), \\ 
    b^l(\bm{z}_l) \sim \mathcal{GP}(0, K_b).
  \end{cases}
\end{align}
In general, $\tilde{w}^l(\bm{z}_l, \bm{z}_{l-1})$ are Gaussian random variables with mean 0 and covariance $K\parentheness{\verbar{\hat{\bm{z}}_l - \tilde{\bm{z}}_l}, \verbar{\hat{\bm{z}}_l - \tilde{\bm{z}}_{l-1}}}$ where $\hat{\bm{z}}_l$ and $\tilde{\bm{z}}_l$ are positions of neurons in the $l$-th layer and $\tilde{\bm{z}}_{l-1}$ is position of neuron in the $(l-1)$-th layer.
We define the covariance by the difference between two positions, e.g., $\hat{\bm{z}}_l - \tilde{\bm{z}}_l$.
This is rational because it makes the neural field have translational symmetry, which is biologically plausible.
For the sake of simplicity, we assume that $\tilde{w}^l(\hat{\bm{z}}_l, \hat{\bm{z}}_{l-1})$ and $\tilde{w}^l(\tilde{\bm{z}}_l, \tilde{\bm{z}}_{l-1})$ are statistically independent when $\hat{\bm{z}}_l \neq \tilde{\bm{z}}_l$. We leave the translational symmetry within each layer, that is, define the $K$ by a univariate function  $K\parentheness{\verbar{\hat{\bm{z}}_{l-1} - \tilde{\bm{z}}_{l-1}}}$.
\begin{figure}[t]
    \centering
    \includegraphics[width=0.9\linewidth]{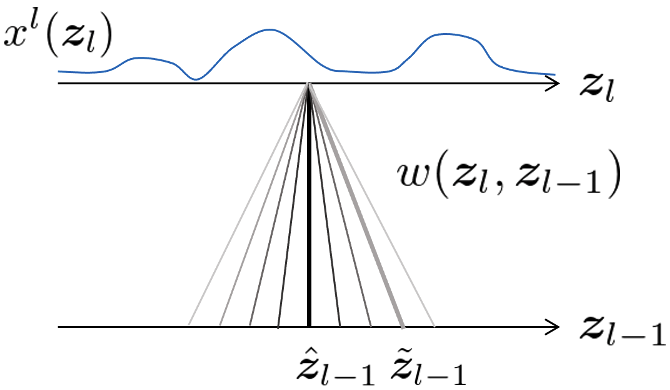}
    \caption{Relation between RNF output and random field.}
    \label{schematic_rnf}
\end{figure}
That is, different neurons are not correlated. Fig. \ref{schematic_rnf} shows the relation between RNF output and random field. Darker lines depict neurons closer to $\hat{\bm{z}}_{l-1}$ which have a higher correlation. The blue line represents the output $x^l(\bm{z}_l)$. We also introduce receptive fields to the random field as will be described in Section 2.2. The connection weights of $\tilde{w}^l(\bm{z}_l, \hat{\bm{z}}_{l-1})$ and $\tilde{w}^l(\bm{z}_l, \tilde{\bm{z}}_{l-1})$ have covariance given by $K\parentheness{\verbar{\hat{\bm{z}}_{l-1} - \tilde{\bm{z}}_{l-1}}}$. A simple example is Gaussian
\begin{align}
    K\parentheness{\verbar{\hat{\bm{z}}_{l-1} - \tilde{\bm{z}}_{l-1}}} = \exp\parentheness{-\dfrac{\verbar{\hat{\bm{z}}_{l-1} - \tilde{\bm{z}}_{l-1}}^2}{2\sigma^2}}.
\end{align}
In addition, as a covariance function, we will also use a Mat\'ern kernel and an RBF kernel on the $d$-dimensional torus $\TT^d$. Their detailed formulations are described in Appendix \ref{form_mat_and_rbf}.

Since the Mat\'ern kernel is a parameterized family of distributions that covers a variety of distributions from Gaussian to Student $t$ and Cauchy, it is convenient to examine the effect of different strengths of correlation in the receptive field. The Mat\'ern kernel is preferable because the corresponding RKHS is a Sobolev space, where the parameter corresponds to the smoothness. Namely, the parameter controls the complexity of the hypothesis class and thus the RKHS is a natural hypothesis space.

\subsection{Neural fields with receptive fields}
Let $r^l(\bm{z}_l, \bm{z}_{l-1})$ denote the receptive field of a neuron at position $\bm{z}_l$ of the $l$th layer such that $r^l(\bm{z}_l, \bm{z}_{l-1})=0$ when neurons at $\bm{z}_{l-1}$ of the $(l-1)$-th layer are outside the receptive field of the neuron at $\bm{z}_l$. We introduce the connection weights as $w(\bm{z}_l, \bm{z}_{l-1}) = \tilde{w}(\bm{z}_l, \bm{z}_{l-1})r^l(\bm{z}_l, \bm{z}_{l-1})$ by using the $\tilde{w}$ of the Gaussian field of (\ref{wab}). In addition to the Gaussian receptive field, we also consider the Mexican-hat-type receptive field, which is a typical feature of the receptive fields in the biological brain \cite{marr1982vision} and also image processing. We define the receptive fields as in Tab. \ref{RF}.
\begin{table}[t]
  \caption{Receptive fields.}
  \label{RF}
  \centering
  \scalebox{1.0}{
  \begin{tabular}{ll}
    \toprule
    Name     &     Definition  \\
    \midrule
    Gaussian filter     &
    $r\left(\bm{z}_{l}, \bm{z}_{l-1}\right)=\exp \left(-\dfrac{\left|\bm{z}_{l}-\bm{z}_{l-1}\right|^{2}}{2 \sigma_{r}^{2}}\right)$  \\
    Mexican hat     &
    $r\left(\bm{z}_{l}, \bm{z}_{l-1}\right)=\dfrac{2}{\sqrt{3 \sigma_r} \pi^{1 / 4}}\left(1-\left(\dfrac{\left|\bm{z}_{l}-\bm{z}_{l-1}\right|}{\sigma_r}\right)^{2}\right) \exp\left(-\dfrac{\left|\bm{z}_{l}-\bm{z}_{l-1}\right|^{2}}{2 \sigma_r^{2}}\right)$ \\
    \bottomrule
  \end{tabular}
  }
\end{table}

\subsection{Convolutional neural networks}
Our model can be easily applied to convolutional neural networks (CNN). From the canonical representation of the Gaussian process \cite{hida1960canonical}, correlated Gaussian fields $w \sim \mathcal{GP}(0, K)$ with covariance function $K$ can be expressed using an appropriate spectral density $f$ as
\begin{align}
    w(\bm{t})\sqrt{\dd\bm{t}} = \int_{\mathbb{T}^d} f(\bm{t}-\bm{s}) \dd B(\bm{s}), \label{canrep}
\end{align}
where $\bm{t} \in \TT^d$ is a vector and $B(\bm{s})$ is a $d$-dimensional Wiener process \cite{hida1960canonical}. Using Eq.(\ref{canrep}), the $d$-RNF can be described as
\begin{align}
    h(\bm{u}) &= \int_{\mathbb{T}^d} w(\bm{t}-\bm{u})x(\bm{t})\sqrt{\dd\bm{t}} = \int_{\mathbb{T}^d} (f \star x)(\bm{1}_d-\bm{s}, \bm{u})\dd B(\bm{s}),
    \label{sec2-1}
\end{align}
where $\bm{1}_d$ is a $d$-dimensional one vector $(1,1,\ldots,1) \in \RR^d$, a periodic boundary condition $x(\bm{t})=x(\bm{t}+\bm{1}_d)$ is assumed, and
\begin{align}
    f \star x(\bm{s}, \bm{u}) \coloneqq \int_{\mathbb{T}^d} f(\bm{t}, \bm{u})x(\bm{t}+\bm{s}) \dd\bm{t}.
    \label{sec2-2}
\end{align}
In this case, the effect of $K$ is aggregated to $f \star x$, which is formally a CNN. That is, the analysis of the $d$-RNF reduces to the analysis of the CNN. However, white noise is not differentiable everywhere. Thus, the convolution is not well-defined. In this paper, we consider the justification for the use of white noise to be the generalized functional theory of infinite variables (see Appendix \ref{whitenoise_convolve}).
\section{Neural Tangent Kernel}
In this section, we describe the neural tangent kernel (NTK), which can capture the learning behavior in a fully connected neural network (FCNN). For input $\bm{x} \in \mathbb{R}^m$, let $f_t(\cdot, \bm{\theta}^t):\mathbb{R}^m \rightarrow \mathbb{R}$ be the output of the neural network at training step $t$, where all the parameters of the network are summarized to $\bm{\theta}^t \in \mathbb{R}^{P} \ (P=\sum_{l=0}^{L-1}(n_{l}+1)n_{l+1})$. For a given training dataset $\{(\bm{x}_i, y_i)\}_{i=1}^N$, supervised learning is carried out to minimize the mean squared error (MSE) loss. For the two inputs $\bm{x}_i$ and $\bm{x}_j$, the NTK $\hat{\Theta}_t(\cdot, \cdot): \mathbb{R}^m \times \mathbb{R}^m \rightarrow \mathbb{R}$ at training step $t$ is defined by
\begin{align}
    \hat{\Theta}_t(\bm{x}_i, \bm{x}_j) \coloneqq \sum_{k=1}^P \dfrac{\partial f_t(\bm{x}_i, \bm{\theta}^t)}{\partial \theta_k^t}\dfrac{\partial f_t(\bm{x}_j, \bm{\theta}^t)}{\partial \theta_k^t}.
\end{align}
The details of update rule for NTK are described in the Appendix \ref{update_rule_for_ntk}. It is known that NTK at any training step $t$ can be approximated by the initial random kernel $\hat{\Theta}_0$, provided the width of the network is sufficiently large \cite{jacot2018neural, lee2019wide}. If $f_t(\mathcal{X})$ is vec$\parentheness{f_t(\bm{x}_1),\ldots,f_t(\bm{x}_N)} \in \mathbb{R}^{N}$, the vector of the concatenated outputs of the network, then NTK $\hat{\Theta}_t$ is an $N \times N$ matrix.

The first-order approximation of the output $f_t(\bm{x}_i, \bm{\theta}^t)$ of the network using Taylor expansion at $\bm{\theta}^0$ gives the following equation:
\begin{align}
    f_t(\bm{x}_i, \bm{\theta}^t) \approx f_0(\bm{x}_i, \bm{\theta}^0) + \iprod{\bm{\theta}^t-\bm{\theta}^0, \frac{\partial f_0(\bm{x}_i, \bm{\theta}^0)}{\partial \bm{\theta}^t}}. \label{taylor_expansion}
\end{align}
Denoting $\mathcal{Y}=(y_1,\ldots, y_N)^\top \in \mathbb{R}^N$, we can linearize the above equation into the following equation using NTK $\hat{\Theta}_0$ in a sufficiently over-parameterized case, i.e., $P \gg N$:
\begin{align}
    f_{t}(\bm{x}') = f_{0}\left(\boldsymbol{x}^{\prime}\right)-\hat{\Theta}_{0}\parentheness{\boldsymbol{x}', \mathcal{X}} \hat{\Theta}_{0}^{-1}\parentheness{I-\exp\parentheness{-\eta N^{-1} \hat{\Theta}_{0}t}}\left(f_{0}(\mathcal{X})-\mathcal{Y}\right). \label{NTK_test_output}
\end{align}
Setting this learning rate $\eta$ appropriately, output dynamics are perfectly fit with those by backpropagation.

\cite{bietti2019inductive} attempted to control various RKHS functions by investigating their smoothness and stability with respect to the deformation and translation of NTK mappings for multilayer networks and CNNs. The average relative distance from the reference image $\bm{x}$ to a deformed image $\bm{x}^{\prime}$ is obtained by applying the following equation:
\begin{align}
    \frac{1}{\verbar{S}} \sum_{\bm{x}' \in S} \frac{\left\|\Phi\left(\bm{x}'\right)-\Phi(\bm{x})\right\|_{\mathcal{H}}}{\|\Phi(\bm{x})\|_{\mathcal{H}}} = \frac{1}{|S|} \sum_{\bm{x}' \in S} \frac{\sqrt{\hat{\Theta}_{0}(\bm{x}, \bm{x})+\hat{\Theta}_{0}\left(\bm{x}', \bm{x}'\right)-2 \hat{\Theta}_{0}\left(\bm{x}, \bm{x}'\right)}}{\sqrt{\hat{\Theta}_{0}(\bm{x}, \bm{x})}},
\end{align}
where $S$ is a set of images, $\mathcal{H}$ is the RKHS associated with the kernel $\hat{\Theta}_{0}$, and $\Phi$ is the kernel mapping. 
\section{Experimental Setup}
\label{experimental_setup}
In this section, we describe the setup for preliminary experiments. The code for reproducing our experiments is found in \url{https://github.com/kwignb/RandomNeuralField}. We used three hidden layer RNF models with an ReLU activation function, and the layer width was 2,048. Stochastic gradient descent (SGD) was used as the optimization method, and the learning rate was $\eta = 2\lambda_{\max}(\hat{\Theta}_0)^{-1}$, where $\lambda_{\max}(\hat{\Theta}_0)$ was the maximum eigenvalue of the NTK. The number of learning epochs was $1,000,000$ in the discrete time steps. To reduce the computational cost, 80 training and 20 validation samples were randomly chosen from the MNIST dataset \cite{lecun1998gradient}. Instead of solving a classification task, models were trained to classify digits using regression. For the labels of classes $c \in \curly{1,\ldots,10}$, we defined $y=-0.1 \cdot \mathbf{1}+\boldsymbol{e}_{c}$, where $\boldsymbol{e}_{c}=(0, \ldots, 1, \ldots, 0) \in \mathbb{R}^{|c|}$ \cite{novak2018bayesian} is the vector in which the $c$th component is $1$ and the other components are $0$. To investigate the robustness of the RNF model to perturbations, we conducted experiments using the infinite MNIST dataset \cite{loosli2007traning}.

\subsection{Discretization}
\label{subs_disc}
For numerical simulations, we define a discretized RNF model as:
\begin{align}
    \begin{cases}
        h^l &= x^{l-1} W^{l} + \bm{b}^l \\
        x^l &= \varphi\left(h^l\right)
    \end{cases}
    ~\text{and}~
    \begin{cases}
        W_{ij}^l &= \frac{\sigma_w}{\sqrt{n_l}}r_{ij}^l \tilde{W}_{ij}^{l-1}, \\
        b_j^l &= \sigma_b \beta_j^l
    \end{cases}
\end{align}
where $\varphi$ is the ReLU activation function, $W^l \in \mathbb{R}^{n_{l-1} \times n_l}$ and $\bm{b}^l \in \mathbb{R}^{n_l}$ are weight matrix and bias vectors, respectively, $r_{ij}^l$ is the receptive field, and $\beta_j^l$ is generated from the standard Gaussian distribution $\mathcal{N}(0,1)$ at initialization. The $j$th vector $\tilde{\bm{w}}^{l}_j$ is generated as:
\begin{align}
    \tilde{\bm{w}}_j^{l} \sim \mathcal{N}(\bm{0}_{n_{l}}, \Sigma^{l}), \quad \Sigma_{ii'}^{l} = \exp\left(-\frac{|i-i'|^2}{2(\sigma_{s_{l}} n_{l})^2}\right)
\end{align}
where $\bm{0}_{n_{l-1}}$ is an $n_{l-1}$-dimensional zero vector. This is the discretization for the Gaussian kernel. See Appendix \ref{disc_and_cor} for the discretization in the case of the Mat\'ern kernel. The receptive fields in the discretized case are shown in Tab. \ref{dRF}.

\begin{table}[t]
  \caption{Receptive fields in the discretized case.}
  \label{dRF}
  \centering
  \scalebox{1.0}{
  \begin{tabular}{ll}
    \toprule
    Name     &     Definition  \\
    \midrule
    Gaussian filter     &
    $r_{ij}^l = \exp\left(-\dfrac{1}{2}\left|\dfrac{1}{\sigma_r}\left(\dfrac{i}{n_{l}}-\dfrac{j}{n_{l-1}}\right)\right|^2\right), \ i = 1,\ldots,n_l, \ j = 1,\ldots, n_{l-1}$  \\
    Mexican hat     &
    $r_{i j}^{l}=\dfrac{2}{\sqrt{3 \sigma_{r}} \pi^{1 / 4}}\left(1-\left|\dfrac{1}{\sigma_{r}}\left(\dfrac{i}{n_{l}}-\dfrac{j}{n_{l-1}}\right)\right|^{2}\right) \exp \left(-\dfrac{1}{2}\left|\dfrac{1}{\sigma_{r}}\left(\dfrac{i}{n_{l}}-\dfrac{j}{n_{l-1}}\right)\right|^{2}\right)$ \\
    \bottomrule
  \end{tabular}
  }
\end{table}

\subsection{Correlations}
\label{subs_corr}
To generate a sample $\tilde{W}^l$ of the Gaussian random field with mean $0$ and covariance $\Sigma$, we use the following proposition.
\begin{prop}
Suppose $Y=(Y_1,\dots,Y_d)^{\textsf{T}}\sim \mathcal{N}(\bm{0}_d,I_d)$. For an arbitrary nondegenerate matrix $A\in\mathbb{R}^{d \times d}$, let $X := AY$. Then, $X \sim \mathcal{N}(\bm{0}_d, AA^{\textsf{T}})$.
\end{prop}
First, we calculate the decomposition $\Sigma^{l-1} = A^{l-1}A^{l-1\textsf{T}}$, which is explained later, using the nondegenerate matrix $A^{l-1}$.
Next, let $\omega^{l-1}_i\sim \mathcal{N}(\bm{0}_{n_{l-1}},I_{n_{l-1}})$ and $\tilde{W}^l := \frac{\sigma_w}{\sqrt{n_l}}R^l \circ \left( A^{l-1}\Omega^l \right),$ where $\circ$ is the Hadamard product, $R^l_{ij}=r^l_{ij}$, and $\Omega^l=(\omega_1^l,\dots,\omega_{n_l}^l)\in\mathbb{R}^{n_{l-1}  \times  n_l}$. Then, by the proposition, $\tilde{W}^l$ is a $d$-dimensional realization of the random field.

We remark that the decomposition of $\Sigma^{l-1}$ is not unique. The Cholesky decomposition is a versatile method for an arbitrary positive definite matrix. However, in implementation, the numerical computation of this decomposition can be time-consuming because $\Sigma^{l-1}$ tends to have a large dimension. Therefore, in this study, we prepare closed-form formulas to compute $A^{l-1}$ by using Fourier calculus. See Appendix \ref{disc_and_cor} for concrete equations.

\subsection{Considered models}
\label{subs_models}
We considered five models, Models 1--5, with different structures of initial weights (Tab. \ref{tableofmodels}). In this preliminary study, only the first layer was a neural field with receptive fields and correlated connections. For other layers, the receptive fields were set to be sufficiently wide, and the correlations were set to zero.

Model 1 was a basic RNF. Model 2 acquired the frequency selectivity and translational invariance of CNNs by using random weights and pooling in the second layer \cite{Saxe2011Randomweights}. Model 3 was prepared for mimicking the visual cortex. We are interested in how lateral inhibition functions in a feedforward network work. Visual information is processed as retina $\to$ LGN (lateral geniculate nucleus) $\to$ V1 (primary visual cortex). Mexican-hat-type center-surround inhibitory connections are abundant in an LGN, and a Gabor filter with directionally selective inhibition plays the main role in V1. A linear combination of Mexican-hat-type filters acts as a Gabor filter. NTK cannot extract features, so a hand-crafted filter is used in the first hidden layer to efficiently embed an effective basis that represents images into a kernel.

A fourth model, Model 4, which had a Mat\'ern kernel, was also prepared. The Mat\'ern kernel has a smoothness parameter $\nu$ that controls the size of the RKHS. The Laplacian kernel can be produced by letting $\nu=0.5$, or the Gaussian kernel can be obtained by letting $\nu=\infty$. The supervised learning behavior of RNF models was compared with that of the NTK vanilla model, which is a three-layer network with NTK parameterization (Model 5). 

We summarize the evaluated model architectures in Tab. \ref{tableofmodels}, and examples of the generated initial weights are presented in Fig. \ref{Initial_weights} in Appendix \ref{weight_matrix_vis}. A fully connected layer is denoted as FC layer, where the weights are assumed to be generated independently from the same distribution as in NTK parameterization. When a receptive field is added, it is denoted as (\textit{receptive field}) FC layer, and when the generation method is changed, it is denoted as FC layer (\textit{generation method}).

\begin{table}[t]
  \caption{Models with three hidden layers: GF=Gaussian filter, MH=Mexican hat, GK=Gaussian kernel, FC= fully connected, and MK=Mat\'ern kernel.}
  \label{tableofmodels}
  \centering
  \begin{tabular}{ll}
    \toprule
    Name     &     Definition  \\
    \midrule
    Model 1 &
    (GF) FC layer (GK) $\to$ FC layer $\to$ FC layer  \\
    Model 2 &
    (GF) FC layer (GK) $\to$ MaxPooling $\to$ FC layer  \\
    Model 3 &
    (MH) FC layer (GK) $\to$ FC layer $\to$ FC layer \\
    Model 4 &
    (GF) FC layer (MK($\nu=0.5$)) $\to$ FC layer $\to$ FC layer \\
    Model 5 &
    FC layer $\to$ FC layer $\to$ FC layer \\
    \bottomrule
  \end{tabular}
\end{table}
\section{Results}
In this section, we give the results of several experiments based on the setup described in Section \ref{experimental_setup}.

\subsection{Random neural fields in NTK regime}
\label{subsection_RNF_NTK_experiments}
In this section, we show by numerical experiments that neural fields follow the NTK regime (Fig. \ref{model1_ntkregime}). We confirm that the other models are also in the NTK regime (see Appendix \ref{other_ntk_regime}). The parameters $\sigma_r$ and $\sigma_s$ used in these experiments were selected from the combination that minimizes the test loss obtained by NTK regression \cite{jacot2018neural} as described below.

\begin{figure}[t]
    \centering
    \includegraphics[width=\linewidth]{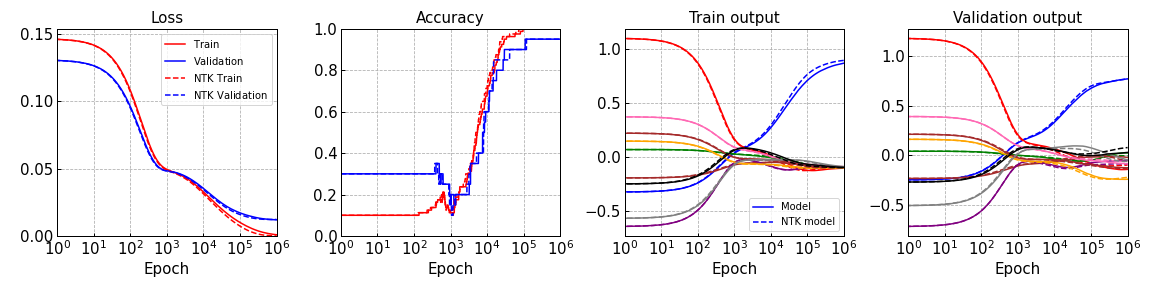}
    \caption{Comparison of SGD/NTK-based methods. Loss/accuracy/output curves for Model 1 with $\sigma_r=0.5$ and $\sigma_s=0.01$. In the right two graphs, each color represents an output value when inputting a specific MNIST digit.}
    \label{model1_ntkregime}
\end{figure}

\subsection{Comparison of test loss values for five models}
Under the assumption that the width of the layers is infinite, the output $\bm{f}^{\star}$ of the fully trained FCNN is equivalent to the output obtained by NTK regression \cite{jacot2018neural}, i.e.,
\begin{align}
\label{ntk_regression}
    \bm{f}^{\star} = \hat{\Theta}_{0}\parentheness{\boldsymbol{x}', \mathcal{X}} \hat{\Theta}_{0}^{-1}\mathcal{Y}.
\end{align}

We investigated the generalization performance in each model by performing NTK regression Eq. (\ref{ntk_regression}) using 800 training and 200 validation samples from the MNIST dataset and by computing the loss and the accuracy. The loss is calculated as:
\begin{align}
    \mathcal{L} = \dfrac{1}{2}\verbar{\bm{f}^{\star} - \bm{y}'}^2,
\end{align}
where $\bm{y}'$ is a label of the test data. The results are shown in Tab. \ref{result_ntkreg}, where the average of five times and their standard deviations are listed. All models were found to be superior to the conventional models, with Model 4 having the smallest loss and the highest accuracy. The search method for parameters $\sigma_r$ and $\sigma_s$ used in the calculations is described in Appendix \ref{grids_ntkreg}.
\begin{table}[t]
    \caption{List of test loss values and test accuracy obtained by NTK regression for five models.}
    \label{result_ntkreg}
    \centering
    \scalebox{0.72}{
    \begin{tabular}{@{}cccccc@{}}
    \toprule
    & \begin{tabular}{c}
         Model 1  \\
         ($\sigma_r=0.5, \sigma_s=0.01$)
    \end{tabular} & \begin{tabular}{c}
         Model 2  \\
         ($\sigma_r=0.5, \sigma_s=0.01$)
    \end{tabular} & \begin{tabular}{c}
         Model 3  \\
         ($\sigma_r=0.01, \sigma_s=0.01$)
    \end{tabular} & \begin{tabular}{c}
         Model 4  \\
         ($\sigma_r=0.5, \sigma_s=0.01$)
    \end{tabular} & Model 5 \\ \midrule
    test loss & $0.0123 \pm 0.0007$ & $0.0130 \pm 0.0006$ & $0.0121 \pm 0.0004$ & $0.0118 \pm 0.0004$ & $0.0156 \pm 0.0005$ \\
    test accuracy & $90.60 \pm 1.83\%$ & $89.80 \pm 1.21\%$ & $91.10 \pm 1.39\%$ & $93.00 \pm 1.55\%$ & $88.90 \pm 1.16\%$ \\ \bottomrule
    \end{tabular}
    }
\end{table}

\subsection{Stability to perturbations}
\label{subsection_robustness}
\begin{figure}[t]
    \begin{minipage}[b]{0.49\linewidth}
        \centering
        \includegraphics[keepaspectratio, scale=0.25]{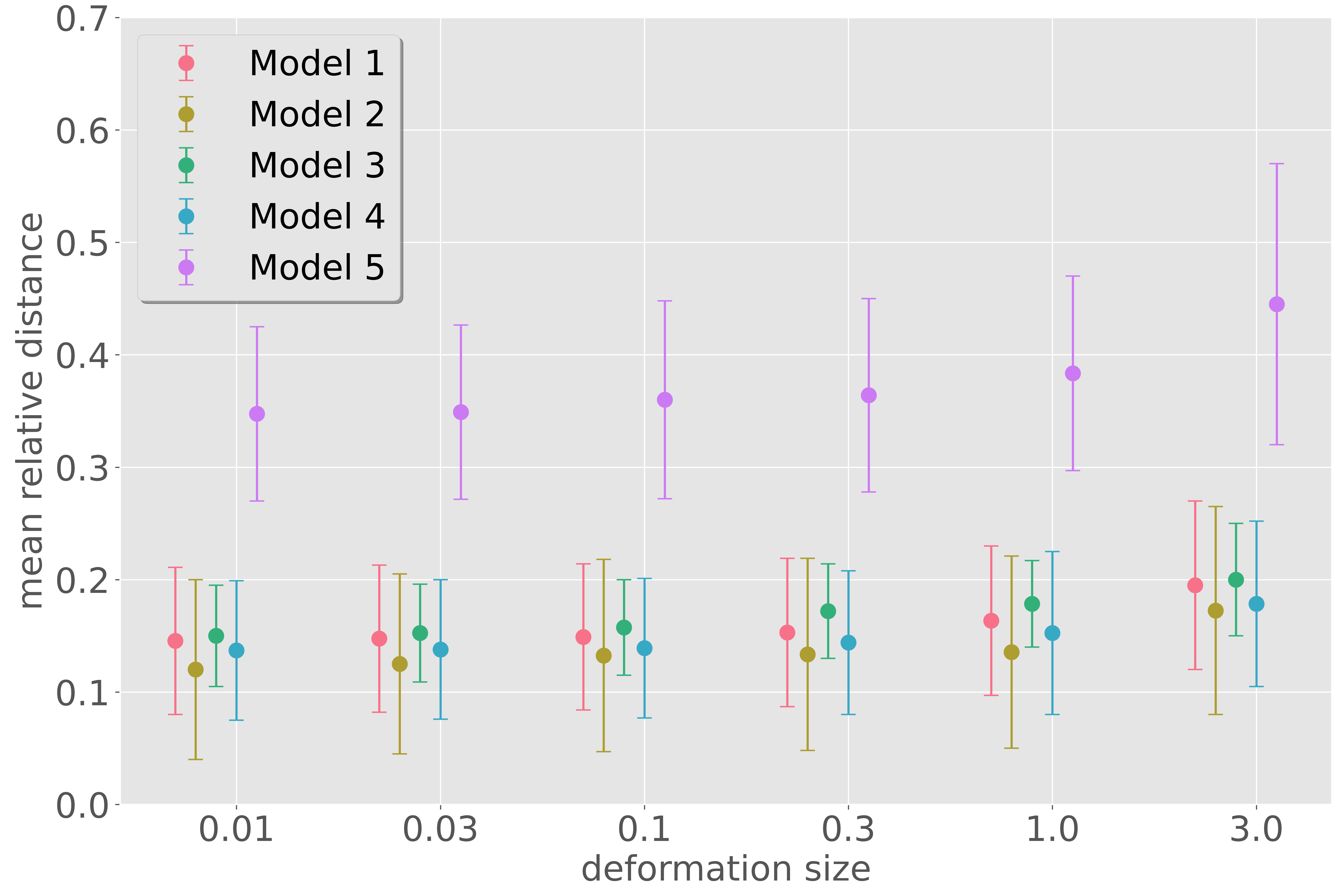}
        \subcaption{Deformations}\label{GeoD}
    \end{minipage}
    \begin{minipage}[b]{0.49\linewidth}
        \centering
        \includegraphics[keepaspectratio, scale=0.25]{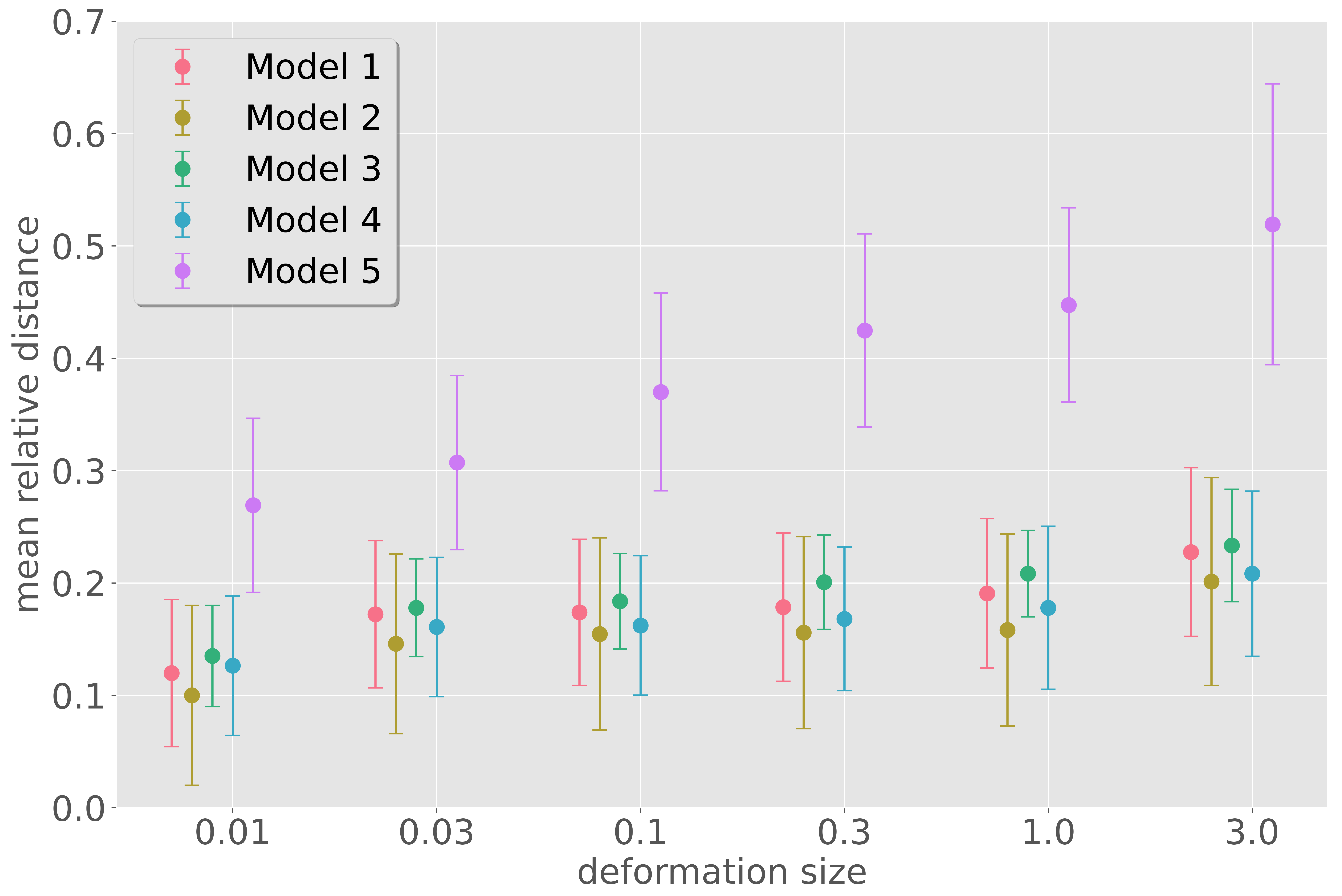}
        \subcaption{Translations and deformations}\label{GeoT}
    \end{minipage}
    \caption{Average relative distances of single digit for different deformations or combinations of translations and deformations from infinite MNIST dataset \cite{loosli2007traning}.}
    \label{Geometry}
\end{figure}

\begin{figure}[t]
    \centering
    \begin{minipage}[b]{0.47\linewidth}
        \centering
        \includegraphics[keepaspectratio, scale=0.3]{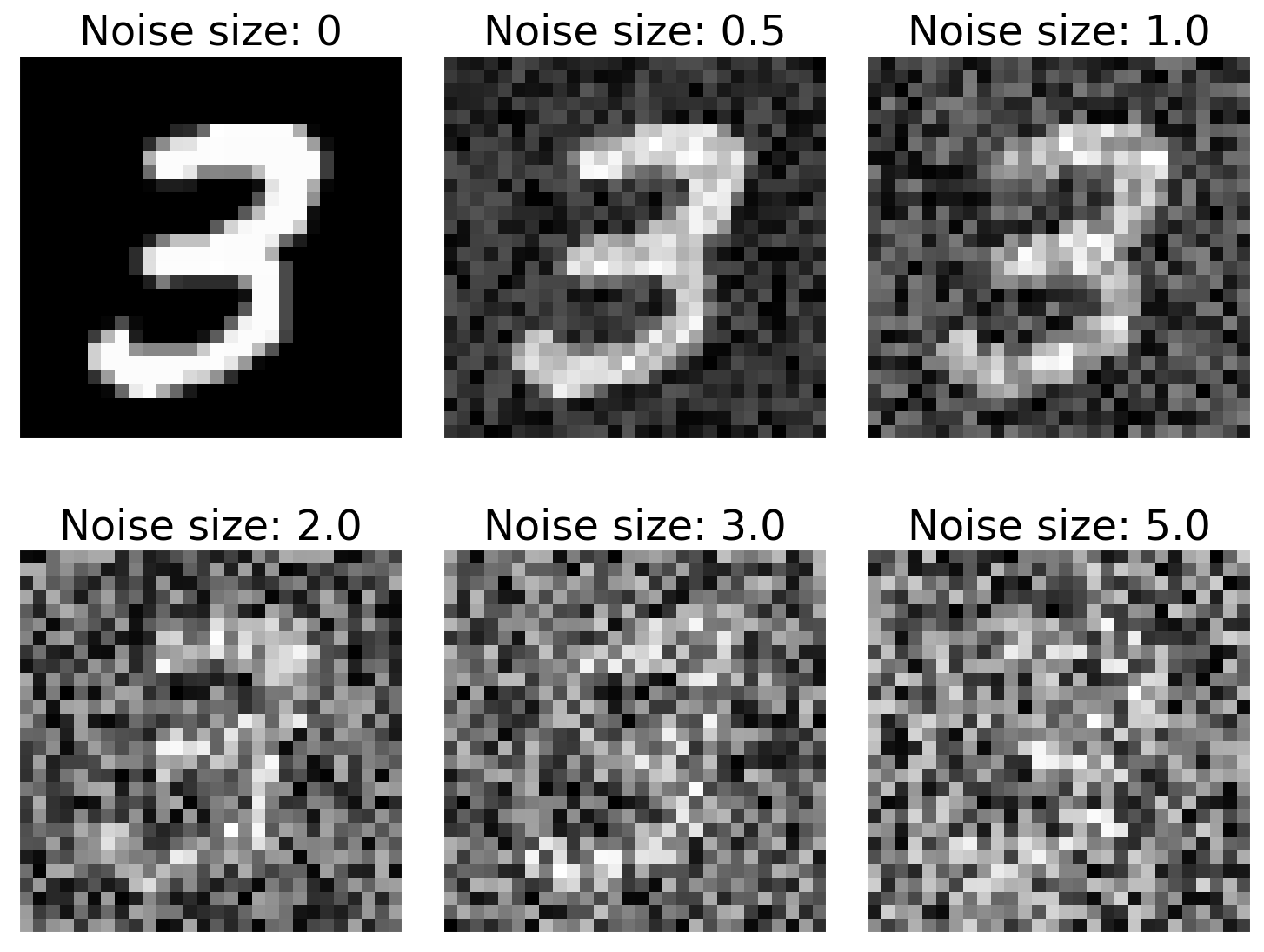}
        \subcaption{Samples from MNIST with different noise levels}
        \label{mnistwithnoise}
    \end{minipage}
    \begin{minipage}[b]{0.49\linewidth}
        \centering
        \includegraphics[keepaspectratio, scale=0.25]{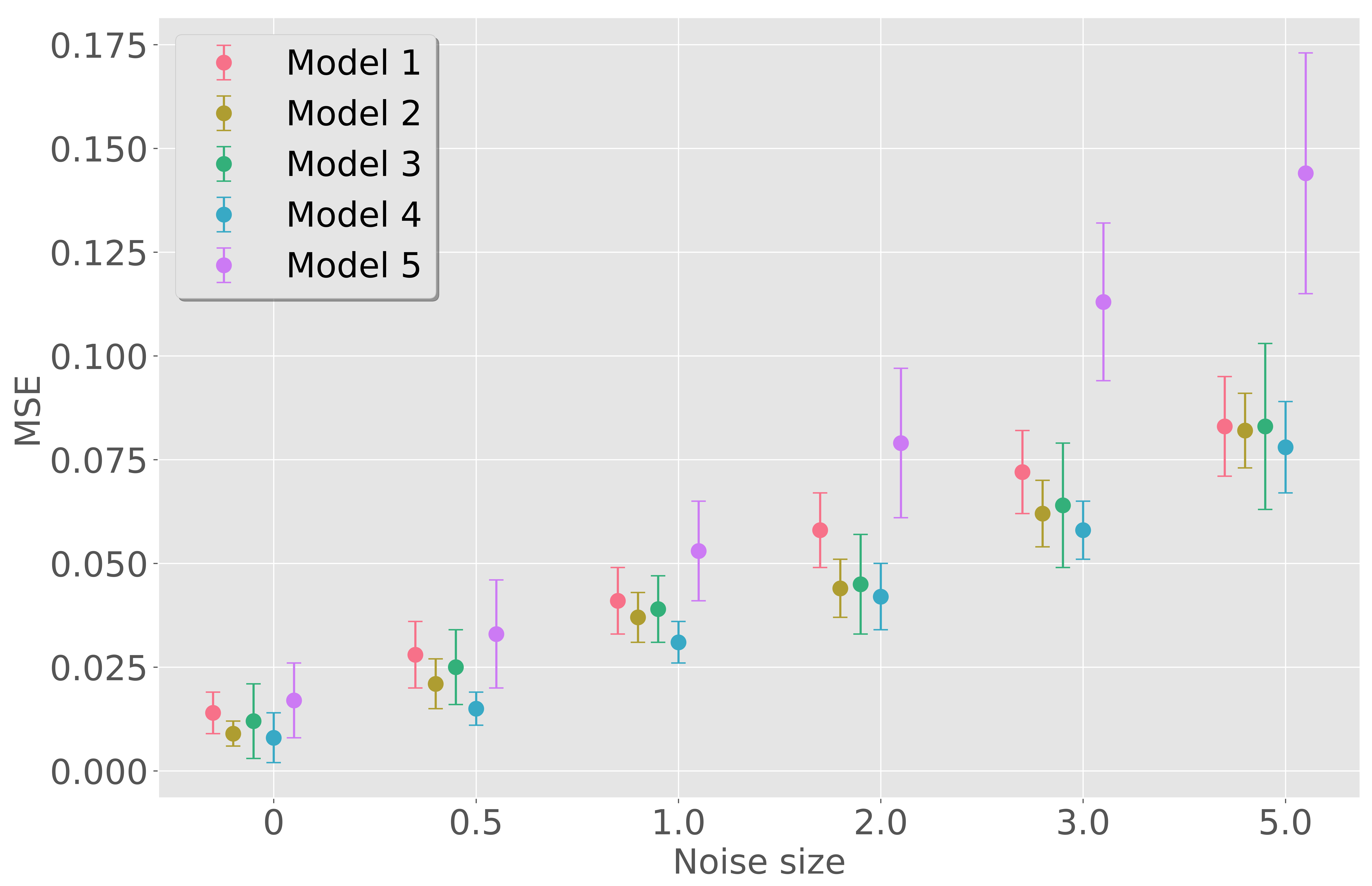}
        \subcaption{MSE loss for different noise levels}
        \label{robustness2noise}
    \end{minipage}
    \caption{Samples from MNIST with stepped noises and the evaluation of robustness of RNFs under different noise levels.}
    \label{fig:grid_search_mrd}
\end{figure}

We numerically investigated the robustness of the neural field models to perturbations. The stability of the kernel mapping representation for our models of neural fields was assessed by following the approach of Bietti and Mairal \cite{bietti2019inductive}. Fig. \ref{Geometry} shows the mean relative distances for five trials and their standard deviations of a single digit for different deformations or combinations of translations and deformations from the infinite MNIST dataset \cite{loosli2007traning}. The model with a single Gaussian filter and a subsequent MaxPooling layer was most robust to the combinations of random translations and deformations. The Mat\'ern kernel model also exhibited robustness to deformations as well as to combinations of translations and deformations. The search method for parameters $\sigma_r$ and $\sigma_s$ used in the calculations is described in Appendix \ref{avg_distance_comb}. We also evaluated the stability of the neural fields to different levels of noise using the average of five trials and its standard deviation (Figs. \ref{mnistwithnoise} and \ref{robustness2noise}). These results are consistent with our expectations that biologically plausible architectures such as receptive fields and correlations perform some regularization. That is, as mentioned in Eqs.(\ref{sec2-1}) and (\ref{sec2-2}), our model can be regarded as a CNN, and according to theoretical analysis by Bietti and Miral \cite{bietti2019inductive}, CNNs are robust to the perturbations.

\section{Conclusion}
We investigated the behaviors of the supervised learning of RNF locally correlated connections and receptive fields in the NTK regime. We confirmed numerically that neural field models are robust to perturbations of training datasets. A possible mechanism behind this is that the associated RKHSs of neural field models are smaller than those of NTK-parameterized vanilla neural networks. Recently, several works have revealed that vanilla deep neural networks with ReLU activation have the same RKHS as the Laplace kernel \cite{chen2020deep}. It would be interesting to theoretically identify the RKHS of our neural field models. This paper provides only preliminary results of the supervised learning of neural fields; nevertheless, we believe that it will inspire future studies on neural fields. In particular, we are interested in the analytical justification of NTK governance in the supervised learning of continuum models with a higher dimensionality ($d\geq2$). Moreover, one can study the effectiveness of the structures in the initial weights further.

\section*{Acknowledgement}
This work was supported by JST PRESTO JPMJPR2125, JSPS KAKENHI 18K18113 and 19K203666 and JST ACT-X JPMJAX190A.

\bibliographystyle{unsrtnat}
\bibliography{references.bib}

\appendix
\section{Mat\'ern Kernel and RBF Kernel on \texorpdfstring{$\TT^d$}{T}}
\label{form_mat_and_rbf}
The formulations of the Mat\'ern kernel and RBF kernel on the $d$-dimensional torus $\TT^d$ are given as follows \cite{borovitskiy2020matern}:
\begin{align}
    K_{\nu}(\bm{z}_{l-1}, \bm{z}'_{l-1}) &= \sum_{\bm{n} \in \ZZ^d} \dfrac{2^{1-\nu}}{C_{\nu}\Gamma(\nu)}\parentheness{\dfrac{\sqrt{2\nu}\verbar{\bm{z}_{l-1}-\bm{z}'_{l-1}+\bm{n}}}{\sigma_{s_{l-1}}}}^{\nu}B_{\nu}\parentheness{\dfrac{\sqrt{2\nu}\verbar{\bm{z}_{l-1}-\bm{z}'_{l-1}+\bm{n}}}{\sigma_{s_{l-1}}}}, \\
    K_{\infty}(\bm{z}_{l-1}, \bm{z}'_{l-1}) &= \sum_{\bm{n} \in \ZZ^d} \dfrac{1}{C_{\infty}}\exp\left(-\dfrac{\left|\bm{z}_{l-1} - \bm{z}'_{l-1}+\bm{n}\right|^2}{2\sigma_{s_{l-1}}^2}\right),
\end{align}
respectively, where $\bm{z}_l$ and $\bm{z}_{l-1}^{\prime}$ are the positions of neurons in the $(l-1)$th layer, $C_{\nu}$ is a constant to ensure $K_{(\cdot)}(\bm{z}, \bm{z})=1$, $B_{\nu}(\cdot)$ is a modified Bessel function of the second kind, $\sigma_{s_{l-1}} \in \mathbb{R}_+$ is a scaling parameter that indicates the degree of the correlation of neurons in the $(l-1)$th layer, and $\nu \in \mathbb{R}_+$ is a parameter that represents the smoothness of a function generated by the Gaussian process.
\section{Justification of White Noise}
\label{whitenoise_convolve}
It is known that white noise is non-differentiable everywhere from the Paley--Wiener--Zygmund theorem. That is, white noise is not Riemann integrable and thus the convolution is not well-defined. In this appendix, we describe how to justify white noise using the generalized function theory of infinite variables.

We consider real-valued function $\xi: \RR \rightarrow \RR$, which satisfies the following conditions.
\begin{itemize}
    \item $\xi$ is a $C^{\infty}$ function.
    \item For any non-negative integers $m$ and $n$, $\underset{x}{\sup} \left|x^m \dfrac{d^n}{dx^n} \xi(x) \right| < \infty$ is satisfied.
\end{itemize}
In this case, $\xi$ is called a rapidly decreasing function on $\RR$, and the whole of the rapidly decreasing function on $\RR$ is denoted by $\mathscr{S}(\RR)$, which is called the Schwartz space. The whole of the continuous linear functionals on $\mathscr{S}(\RR)$ is represented by $\mathscr{S}'(\RR)$, where the elements of $\mathscr{S}'(\RR)$ are Schwartz's generalized functions. We use the standard bilinear form $b(x, \xi)$ on $\mathscr{S}'(\RR) \times \mathscr{S}(\RR)$ for descriptive simplicity. If $f(t)$ is a rapidly decreasing function, then by
\begin{align}
    b(x_f, \xi) = \int_{\RR} f(t)\xi(t) \dd t,
\end{align}
$x_f$ is a continuous linear functional on $\mathscr{S}(\RR)$, that is, it is a generalized function. Since function $f$ defines one generalized function, the Gelfand triple
\begin{align}
    \mathscr{S}(\RR) \subset L^2(\RR) \subset \mathscr{S}'(\RR),
\end{align}
of the inclusion relation in the function space holds.

$\mathscr{S}'(\RR)$ introduces a probability measure, $\mu$, that is uniquely determined by
\begin{align}
    \int_{\mathscr{S}'(\RR)} \exp(ib(x, \xi)) \mu (\dd x) = \exp\left(-\dfrac{b(\xi,\xi)}{2}\right).
\end{align}
This is called the Gaussian measure, and the probability space $(\mathscr{S}'(\RR), \mu)$ is called the Gaussian space. If we fix $\xi \in \mathscr{S}(\RR)$ and set
\begin{align}
    x \longmapsto X_{\xi}(x) = b(x, \xi), \label{eq:gbf}
\end{align}
then $X_{\xi}$ is a function on $\mathscr{S}'(\RR)$; that is, a random variable defined on $(\mathscr{S}'(\RR), \mu)$. Now, $X_{\xi}$ is a Gaussian family with mean 0 and covariance
\begin{align}
    \EE[X_{\xi}X_{\eta}] = b(\xi, \eta), \quad \xi, \eta \in \mathscr{S}(\RR).
\end{align}

We recognize that 
\begin{align}
    X(f) = \int_{\RR} f(t)W(t) \dd t \label{eq:wn}
\end{align}
is a Gaussian random variable for function $f(t)$ that guarantees that $B(t)$ is a Gaussian process. $X(f)$ is a Gaussian family with mean 0 and covariance $\text{Cov}[X(f), X(g)]=b(f, g)$.

Now, let
\begin{align}
    X_{\xi}(x) = b(x, \xi) = \int_{\RR} x(t)\xi(t) \dd t
\end{align}
be the integral representation of Eq.(\ref{eq:gbf}), and if we compare with Eq.(\ref{eq:wn}), we can represent it as $w(t) = x(t) = b(x,\delta_t)$. 

The whole sequence of functions $\phi = (f_n)_{n=0}^{\infty}, f_n \in L^2(\RR^n)_{\text{sym}}$ satisfying
\begin{align}
    \|\phi\|^2 = \sum_{n=0}^{\infty} n! |f_n|^2 < \infty
\end{align}
forms a Hilbert space. This is called the Boson Fock space on $L^2(\RR)$, which is denoted by $\Gamma(L^2(\RR))$. For any $\xi \in \mathscr{S}(\RR)$, the corresponding $\varphi_{\xi} \in L^2(\mathscr{S}'(\RR), \mu)$, and $\left(1, \xi, \xi^{\otimes2}/2!,\ldots,\xi^{\otimes n}/n!\right) \in \Gamma(L^2(\RR))$ defined in
\begin{align}
    \varphi_{\xi}(x) = \exp\left(\dfrac{b(x,\xi)-b(\xi,\xi)}{2}\right), \quad x \in \mathscr{S}'(\RR),
\end{align}
we obtain the unitary isomorphism $L^2(\mathscr{S}'(\RR), \mu) \cong \Gamma(L^2(\RR))$ with the Wiener--It\^o--Segal isomorphism.

If we apply the Wiener--It\^o--Segal isomorphism to the Gelfand triplet
\begin{align}
    \Gamma(\mathscr{S}(\RR)) \subset \Gamma(L^2(\RR)) \subset \Gamma(\mathscr{S}'(\RR)),
\end{align}
we have the inclusion relation
\begin{align}
    \mathscr{W} \subset L^2(\mathscr{S}'(\RR), \mu) \subset \mathscr{W}'
\end{align}
of the function space on the Gaussian space. This is a Gelfand triplet in Gaussian space; the elements of $\mathscr{W}'$ are generalized functions in Gaussian space, and this becomes the generalized function of white noise. By referring to it as white noise, the following holds.
\begin{thm}
    The Wiener process $t \mapsto B(t) \in \mathscr{W}'$ and white noise $t \mapsto W(t) \in \mathscr{W}'$ are both $C^{\infty}$, and
    \begin{align}
        \dfrac{\dd}{\dd t} B(t) = W(t)
    \end{align}
    holds on $\mathscr{W}'$. \label{thm:wn_dif}
\end{thm}
This theorem shows that white noise is differentiable everywhere and that the convolution is well-defined.
\section{Update Rule for Neural Tangent Kernel}
\label{update_rule_for_ntk}
Let $\mathcal{D} \subseteq \mathbb{R}^{n_0} \times \mathbb{R}^{k}$ be a set of dataset, $\mathcal{X}=\{\bm{x} \ | \ (\bm{x},\bm{y}) \in \mathcal{D}\}$ and $\mathcal{Y}=\{\bm{y} \ | \ (\bm{x},\bm{y}) \in \mathcal{D}\}$ as input data and labels, respectively. We assume that the middle layer is $L$-layer, the width of each layer is $n_l$ $(l=1,\ldots,L)$, and the width of the output layer (number of classes) is $n_{L+1}=k$. For the input $\bm{x} \in \mathbb{R}^{n_0}$, let $h^{l}(\bm{x}), x^{l}(\bm{x}) \in \mathbb{R}^{n_l}$ be the pre-activation function and post-activation function. Then, we denote the definition of the recurrence relation of the neural network by
\begin{align}
    \begin{dcases}
        h^{l+1} = x^{l}\bm{W}^{l+1} + \bm{b}^{l+1} \\
        x^{l+1} = \varphi(h^{l+1})
    \end{dcases}
    \ \mbox{and} \  
    \begin{dcases}
        W_{ij}^{l} = \dfrac{\sigma_{\omega}}{\sqrt{n_l}}\omega_{ij}^{l} \\
        b_{j}^{l} = \sigma_b \beta_{j}^{l}
    \end{dcases},
\end{align}

where $\varphi$ is a point-wise activation function, $\bm{W}^{l+1} \in \mathbb{R}^{n_{l} \times n_{l+1}}$ and $\bm{b}^{l+1} \in \mathbb{R}^{n_{l+1}}$ represent the weight matrix and bias vector, and $\omega_{ij}^{l},\beta_{j }^{l}$ are initialized following the standard Gaussian distribution $\mathcal{N}(0, 1)$.

We define the parameter vector $\bm{\theta}^l \in \mathbb{R}^{(n_{l-1}+1)n_l}$ for each layer $L$ and the parameter vector for the whole network $\bm{\theta} \in \mathbb{R}^P \ (P=\sum_{l=0}^{L-1}(n_{l}+1)n_{l+1})$ is defined as follows:
\begin{align}
    \bm{\theta}^{l} \equiv \mbox{vec}\left(\{\bm{W}^{l}, \bm{b}^{l}\}\right), \quad \bm{\theta} = \mbox{vec}\left(\bigcup_{l=1}^L \bm{\theta}^l\right).
\end{align}

Let $\hat{\bm{y}}$ be the predicted value by the neural network, the loss function is $\ell(\hat{\bm{y}}, \bm{y}):\mathbb{R}^{k} \times \mathbb{R}^k \rightarrow \mathbb{R}$. In supervised learning, minimize the empirical loss $\mathcal{L}_t = \sum_{(\bm{x},\bm{y}) \in \mathcal{D}} \ell(f_t(\bm{x},\bm{\theta}), \bm{y})$ by learning $\bm{\theta}$.
The learning is done by changing the parameters in the opposite direction of the gradient in order to reduce the loss function. The learning of the parameters is described below, considering batch learning:
\begin{align}
    \dot{\bm{\theta}}_t = -\eta \dfrac{\partial \mathcal{L}}{\partial \bm{\theta}} = -\eta \nabla_{\bm{\theta}}f_t(\mathcal{X})^{\mathrm{T}}\nabla_{f_t(\mathcal{X})}\mathcal{L}, \label{param1}
\end{align}
where $\eta$ is the learning rate. where $\dot{\bm{\theta}}$ is the time derivative of $\bm{\theta}$ and $\partial/\partial x = \nabla_x$. Also, $\nabla_{f_t(\mathcal{X})}\mathcal{L}$ is the gradient of the loss with respect to the network output $f_t(\mathcal{X})$, $f_t(\mathcal{X}) = \mbox{vec}\parentheness{f_t(\bm{x}_1),\ldots,f_t(\bm{x}_N)} \in \mathbb{R}^{N}$. With respect to the function representing the output $f_t(\mathcal{X})$ obtained by the network, as in the learning of the parameter $\bm{\theta}_t$, we consider the time derivative as follows:
\begin{align}
    \dot{f}_t(\mathcal{X}) = \dfrac{\partial f_t(\mathcal{X})}{\partial t} = \nabla_{\bm{\theta}}f_t(\mathcal{X})\dot{\bm{\theta}_t}.
\end{align}
By Eq.(\ref{param1}),
\begin{align}
    \dot{f}_t(\mathcal{X}) = -\eta\nabla_{\bm{\theta}}f_t(\mathcal{X})\nabla_{\bm{\theta}}f_t(\mathcal{X})^{\mathrm{T}}\nabla_{f_t(\mathcal{X})}\mathcal{L} = -\eta\hat{\Theta}_t(\mathcal{X}, \mathcal{X})\nabla_{f_t(\mathcal{X})}\mathcal{L} \label{func_learning}
\end{align}
holds.

\section{Discretization and Correlations}
\label{disc_and_cor}

\subsection{Proof of Proposition 1 (decomposition of covariance matrix)}
\begin{proof} For any measurable set of Euclidean space $E$,
	\begin{align}
	P( Y \in E ) &= \int_E (2 \pi)^{-d/2} \exp\left( - \dfrac{|\bm{y}|^2}{2} \right) \dd \bm{y} \\
	&= \int_{A(E)}  (2 \pi)^{-d/2} \exp\left( -\dfrac{|A^{-1} \bm{x}|^2}{2} \right) \left|\dfrac{\partial \bm{y}}{\partial \bm{x}}\right| \dd \bm{x} \\
	&= \int_{A(E)}  (2 \pi )^{-d/2} |A|^{-1} \exp\left( -\dfrac{\bm{x}^\top (A A^\top)^{-1} \bm{x}}{2} \right) \dd \bm{x}
	\end{align}
	where we use $|A^{-1} \bm{x}|^2 = \bm{x}^\top (A^{-1})^\top A^{-1} \bm{x}$ and $|\partial \bm{y} / \partial \bm{x}| = |A^{-1}| = |A|^{-1}$. Thus, if we have $\Sigma = A A^\top$ for some $A$, then $X \sim \mathcal{N}(\bm{0}_d, A A^\top)$.
\end{proof}

\subsection{Decomposition of covariance}
Consider the system
\begin{align}
    \displaystyle \Sigma (\bm{x}, \bm{y}) = \int_{\mathbb{R}^d} A(\bm{x},\bm{z})A^{*}(\bm{z},\bm{y})\dd\bm{z} = f(\bm{x}-\bm{y}), \quad \bm{x}, \bm{y}, \bm{z} \in \mathbb{R}^d. \label{assumption_cov}
\end{align}
We assume that $f$ is a real even function and that $A(\bm{x},\bm{y})=g(\bm{x}-\bm{y})$ for some real even function $g$. Then, $A^{*}(\bm{x}, \bm{y})=g^{*}(\bm{x}-\bm{y})=g(\bm{y}-\bm{x})$, and
\begin{align}
    \displaystyle \int_{\mathbb{R}^d}A(\bm{x},\bm{z})A^{*}(\bm{y},\bm{z})\dd\bm{z} &= \int_{\mathbb{R}^d} g(\bm{x}-\bm{z})g(\bm{z}-\bm{y})\dd\bm{z} \nonumber \\
    &= \int_{\mathbb{R}^d}g(\bm{z})g(\bm{y}-\bm{x}+\bm{z})\dd\bm{z} \nonumber \\
    &= \int_{\mathbb{R}^d}g(\bm{z})g((\bm{x}-\bm{y})-\bm{z})\dd\bm{z} \nonumber \\
    &= (g*g)(\bm{x}-\bm{y}). \label{convolution_function}
\end{align}
Thus, Eq.(\ref{assumption_cov}) is reduced to
\begin{align}
    (g*g)(\bm{x}-\bm{y}) = f(\bm{x}-\bm{y}), \quad \bm{x},\bm{y} \in \mathbb{R}^d.
\end{align}
By taking the Fourier transform in $\bm{x}$, we obtain
\begin{align}
    \displaystyle \hat{g}^2(\bm{\xi}) = \hat{f}(\bm{\xi}) = \int_{\mathbb{R}^d}f(\bm{x})\exp(-i\bm{x}\cdot\bm{\xi})\dd\bm{x}, \quad \bm{\xi} \in \mathbb{R}^d, \label{fourie_transform}
\end{align} 
where $\bm{x}\cdot\bm{\xi}$ represents the dot product of $\bm{x}$ and $\bm{\xi}$. Therefore, $g(\bm{x})$ can be represented as
\begin{align}
    \displaystyle g(\bm{x}) = \mathcal{F}^{-1}\left[\sqrt{\hat{f}(\bm{\xi})}\right] = (2\pi)^{-d} \int_{\mathbb{R}^d}\sqrt{\hat{f}(\bm{\xi})}\exp(i\bm{\xi}\cdot\bm{x})\dd\bm{\xi}, \quad \bm{\xi} \in \mathbb{R}^d, \label{g_decomposition_f}
\end{align}
where $\mathcal{F}^{-1}$ is a Fourier inverse transform.

\subsection{Case of Gaussian kernel}
\label{case_gk}
$f(\bm{x})$ and $\hat{f}(\bm{\xi})$ are given as:
\begin{align}
    f(\bm{x}) &= \exp\left(-\dfrac{|\bm{x}|^2}{2\sigma_s^2}\right), \quad \bm{x} \in \mathbb{R}^d \\
    \hat{f}(\bm{\xi}) &= (2\pi)^{d/2}\sigma_s^d \exp\left(-\dfrac{\sigma_s^2|\bm{\xi}|^2}{2}\right), \quad \bm{\xi} \in \mathbb{R}^d. \label{gaussian_fourier}
\end{align}
Substituting Eq.(\ref{gaussian_fourier}) into Eq.(\ref{g_decomposition_f}), we obtain
\begin{align}
    g(\bm{x}) &= \mathcal{F}^{-1}\left[\sqrt{\hat{f}(\bm{\xi})}\right] \\
    &= \mathcal{F}^{-1}\left[(2\pi)^{d/4}\sigma_s^{d/2} \exp\left(-\dfrac{(\sigma_s^2/2)|\bm{\xi}|^2}{2}\right)\right] \\
    &=\dfrac{2^{d/2}}{\sigma_s^{d/2}(2\pi)^{d/4}} \mathcal{F}^{-1}\left[(2\pi)^{d/2}\left(\dfrac{\sigma_s}{\sqrt{2}}\right)^d\exp\left(-\dfrac{(\sigma_s/\sqrt{2})^2|\bm{\xi}|^2}{2}\right)\right] \\
    &= \left(\dfrac{2}{\pi\sigma_s^2}\right)^{d/4}\exp\left(-\dfrac{|\bm{x}|^2}{\sigma_s^2}\right),
\end{align}
and thus,
\begin{align}
    A(\bm{x},\bm{y}) = g(\bm{x}-\bm{y}) = \left(\dfrac{2}{\pi\sigma_s^2}\right)^{d/4}\exp\left(-\dfrac{|\bm{x}-\bm{y}|^2}{\sigma_s^2}\right).
\end{align}
By using white noise $W(\bm{x})$ with $\sigma^2=1$ on $\mathbb{R}^d$ with
\begin{align}
    S(\bm{x}) := (AW(\bm{x})) = \left(\dfrac{2}{\pi\sigma_s^2}\right)^{d/4}\int_{\mathbb{R}^d}\exp\left(-\dfrac{|\bm{x}-\bm{y}|^2}{\sigma_s^2}\right)W(\bm{y})\dd\bm{y},
\end{align}
we obtain
\begin{align}
    &\mathbb{E}[S(\bm{x})] \equiv 0 \\
    &\mathbb{E}[S(\bm{x})S(\bm{y})] = f(\bm{x}-\bm{y}) = \exp\left(-\dfrac{|\bm{x}-\bm{y}|^2}{\sigma_s^2}\right),
\end{align}
and $S(\bm{x})$ is a Gaussian process with $f(\bm{x}-\bm{y})$ as its covariance. Since the model used in this study is $1$-RNF, we consider the case of $d=1$. Let the continuous variables be $x, y \in \mathbb{R}$, and discretize them into discrete variables $i$ and $j$, which are elements of the $n$-dimensional vector. We replace each of them according to the following rules.
\begin{align}
    x_i &:= \dfrac{1}{n}\left(i-\dfrac{n}{2}\right), \quad i \in [n] \\
    y_i &:= \dfrac{1}{n}\left(j-\dfrac{n}{2}\right), \quad j \in [n].
\end{align}
For simplicity of description, the symbols
\begin{align}
    \Delta x &:= (x_{i+1} - x_{i}) = \dfrac{1}{n} \\
    \Delta y &:= (y_{i+1} - y_{i}) = \dfrac{1}{n}
\end{align}
are introduced. Now, the relation
\begin{align}
    \displaystyle f(x-x') = \int_{\mathbb{R}}g(x-y)g(x'-y)\dd y, \quad \forall x,x' \in \mathbb{R}, \label{continous_system}
\end{align}
where
\begin{align}
    f(x) &= \exp\left(-\dfrac{x^2}{2\sigma_s^2}\right) \\ 
    g(x) &= \left(\dfrac{2}{\pi\sigma_s^2}\right)^{1/4}\exp\left(-\dfrac{x^2}{\sigma_s^2}\right),
\end{align}
holds from the discussion of the continuous system. We obtain the decomposition of the covariance matrix $\Sigma=AA^{\mathrm{T}}$ using Eq.(\ref{continous_system}). By using the partitioning quadrature method, we obtain
\begin{align}
    f(x-x') \approx \sum_{j=1}^{n} g(x-y_j)g(x'-y_j)\Delta y, \quad \forall x,x' \in \mathbb{R}.
\end{align}
Here, the covariance matrix $\Sigma$ and the transformation matrix $A$ are defined by $f$ as:
\begin{align}
    \Sigma_{ii'} &= f(x_i-x_{i'}) \\
    &= \exp\left(-\dfrac{|x_i-x_{i'}|^2}{2\sigma_s^2}\right) \\
    &= \exp\left(-\dfrac{|i-i'|^2}{2(\sigma_s/\Delta x)^2}\right) \label{transform_Sigma}\\
    A_{ij} &= g(x_i-y_j)\sqrt{\Delta y} \\
    &= \left(\dfrac{2}{\pi\sigma_s^2}\right)^{1/4}(\Delta y^2)^{1/4}\exp\left(-\dfrac{|x_i-y_j|^2}{\sigma_s^2}\right) \\
    &= \left(\dfrac{2}{\pi(\sigma_s/\Delta y)^2}\right)^{1/4}\exp\left(-\dfrac{|i-j|^2}{(\sigma_s/\Delta y)^2}\right), \label{transform_A}
\end{align}
where $\Delta x = \Delta y$. From Eqs.(\ref{transform_Sigma}) and (\ref{transform_A}), we obtain
\begin{align}
    \Sigma_{ii'} \approx \sum_{j=1}^n A_{ij}A_{i'j}, \quad \forall i,i'.
\end{align}
That is, $\Sigma \approx AA^{\top}$ holds.

\subsection{Case of Mat\'ern kernel}
In the case of the Mat\'ern kernel, the normal Fourier transform of a radial function is not possible. We transform it using the following theorem.
\begin{thm}
    Let $r=\verbar{\bm{x}}$ and $s=\verbar{\bm{\xi}}$, and write $f(\bm{x})=F(r)$ and $\hat{f}(\bm{\xi})=\hat{F}_d(s)$. The radial Fourier transform in $d$-dimensions is given in terms of the Hankel transform by
    \begin{align}
        s^{\frac{d}{2}-1}\hat{F}_d(s) = (2\pi)^{\frac{d}{2}}\int_0^{\infty} J_{\frac{d}{2}-1}(sr)r^{\frac{d}{2}-1}F(r)r\dd r,
    \end{align}
    where $J_{\nu}(\cdot)$ is a Bessel function of the first kind.
\end{thm}
The Mat\'ern kernel is represented by the following equation:
\begin{align}
    f(\bm{x}) = \dfrac{2^{1-\nu}}{\Gamma(\nu)}\left(\dfrac{\sqrt{2\nu}|\bm{x}|}{\theta}\right)^\nu B_{\nu}\left(\dfrac{\sqrt{2\nu}|\bm{x}|}{\theta}\right).
\end{align}
By using this theorem to perform the transformation, we obtain
\begin{align}
    \hat{f}(\bm{\xi}) = \hat{F}_d(s) &= s^{-\frac{d}{2}+1}(2\pi)^{\frac{d}{2}}\int_0^{\infty} J_{\frac{d}{2}-1}(sr)r^{\frac{d}{2}-1}F(r)r\dd r \\
    &= \dfrac{\Gamma\parentheness{\dfrac{d}{2}+\nu}}{\Gamma(\nu)}\dfrac{(2\theta)^d (2\nu)^{\nu}\pi^{\frac{d}{2}}}{\parentheness{s^2 \theta^2 + 2\nu}^{\frac{d}{2}+\nu}}.
\end{align}
When $d>1$, the equations and calculations become very complex; hence, we will consider the case of $d=1$ in accordance with the $1$-RNF of the model used. In the case of $d=1$, we obtain
\begin{align}
    \hat{f}(\xi) &= \dfrac{\Gamma\parentheness{\nu+\frac{1}{2}}}{\Gamma(\nu)}\dfrac{2\theta (2\nu)^{\nu}\sqrt{\pi}}{\parentheness{\xi^2 \theta^2 + 2\nu}^{\nu+\frac{1}{2}}}. \label{matern_d1}
\end{align}
Substituting Eq.(\ref{matern_d1}) into Eq.(\ref{g_decomposition_f}), we obtain
\begin{align}
    g(x) &= \mathcal{F}^{-1}\left[\sqrt{\hat{f}(\xi)}\right] \\
    &= \dfrac{1}{\sqrt{\pi\theta\Gamma(2\nu)}} \parentheness{\dfrac{|x|}{\theta}}^{\nu} B_{\nu}\parentheness{\dfrac{\sqrt{2\nu}|x|}{\theta}},
\end{align}
and thus
\begin{align}
    A(x, y) = g(x-y) = \dfrac{1}{\sqrt{\pi\theta\Gamma(2\nu)}} \parentheness{\dfrac{|x-y|}{\theta}}^{\nu} B_{\nu}\parentheness{\dfrac{\sqrt{2\nu}|x-y|}{\theta}}.
\end{align}
Following the argument in \ref{case_gk}, we obtain
\begin{align}
    \Sigma_{ii'} &= f(x_i-x_{i'}) \\
    &= \dfrac{2^{1-\nu}}{\Gamma(\nu)}\left(\dfrac{\sqrt{2\nu}|x_i - x_{i'}|}{\theta}\right)^\nu B_{\nu}\left(\dfrac{\sqrt{2\nu}|x_i - x_{i'}|}{\theta}\right) \\
    &= \dfrac{2^{1-\nu}}{\Gamma(\nu)}\left(\dfrac{\sqrt{2\nu}|i - i'|\Delta x}{\theta}\right)^\nu B_{\nu}\left(\dfrac{\sqrt{2\nu}|i - i'|\Delta x}{\theta}\right) \\
    A_{ij} &= g(x_i - y_j)\sqrt{\Delta y} \\
    &= \dfrac{1}{\sqrt{\pi\theta\Gamma(2\nu)}} \parentheness{\dfrac{|x_i - y_j|}{\theta}}^{\nu} B_{\nu}\parentheness{\dfrac{\sqrt{2\nu}|x_i - y_j|}{\theta}} \sqrt{\Delta y} \\
    &= \sqrt{\dfrac{\Delta y}{\pi\theta\Gamma(2\nu)}} \parentheness{\dfrac{|(i-j)\Delta y|}{\theta}}^{\nu} B_{\nu}\parentheness{\dfrac{\sqrt{2\nu}|(i-j)\Delta y|}{\theta}}.
\end{align}
\section{Further Experimental Details}
\label{experiment_details}
\subsection{Visualization of initial weight matrix}
\label{weight_matrix_vis}
\begin{figure}[t]
\captionsetup[subfigure]{justification=centering}
  \begin{minipage}[b]{0.33\linewidth}
    \centering
    \includegraphics[keepaspectratio, scale=0.18]{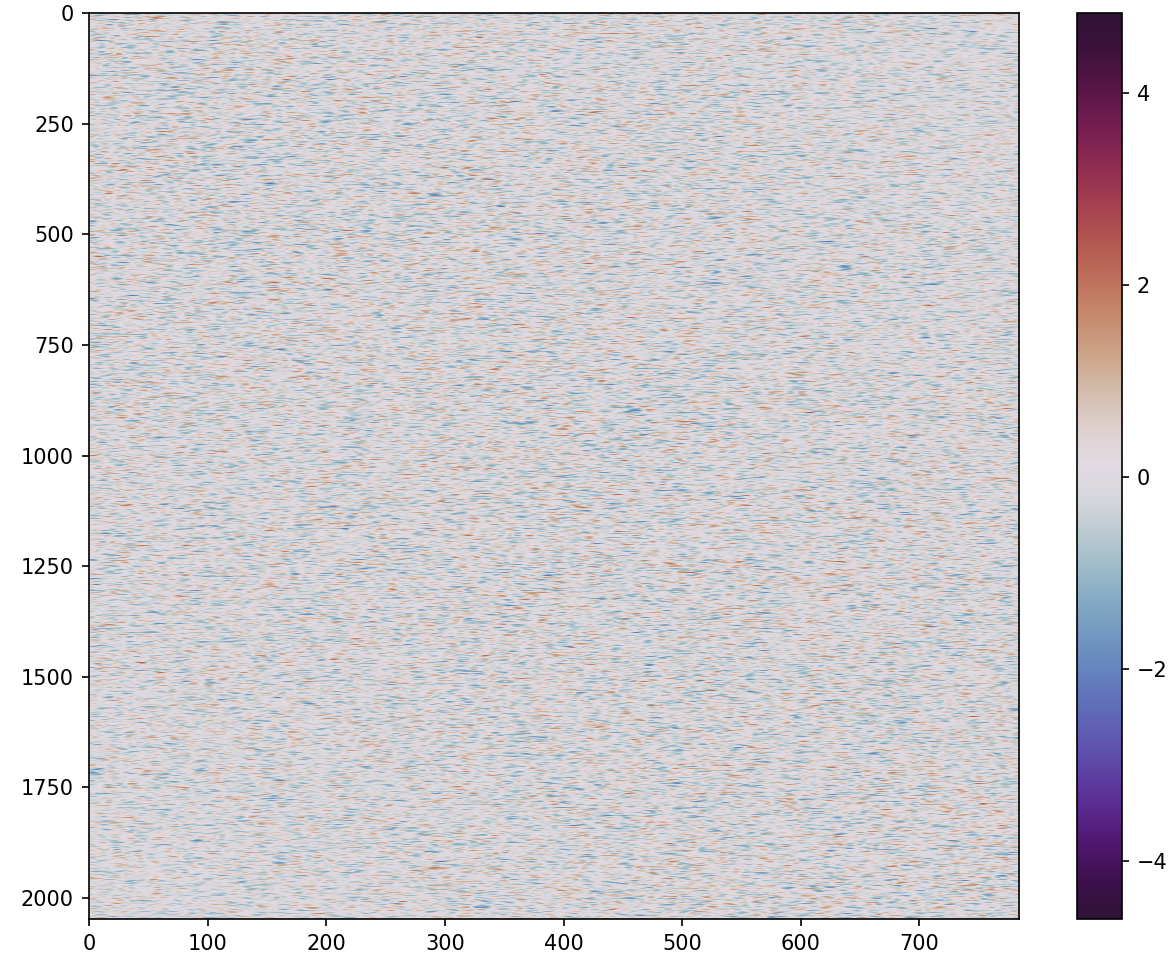}
    \subcaption{Model 1 with \\ $\sigma_r=1.0$, $\sigma_s=0.01$}
  \end{minipage}
  \begin{minipage}[b]{0.33\linewidth}
    \centering
    \includegraphics[keepaspectratio, scale=0.18]{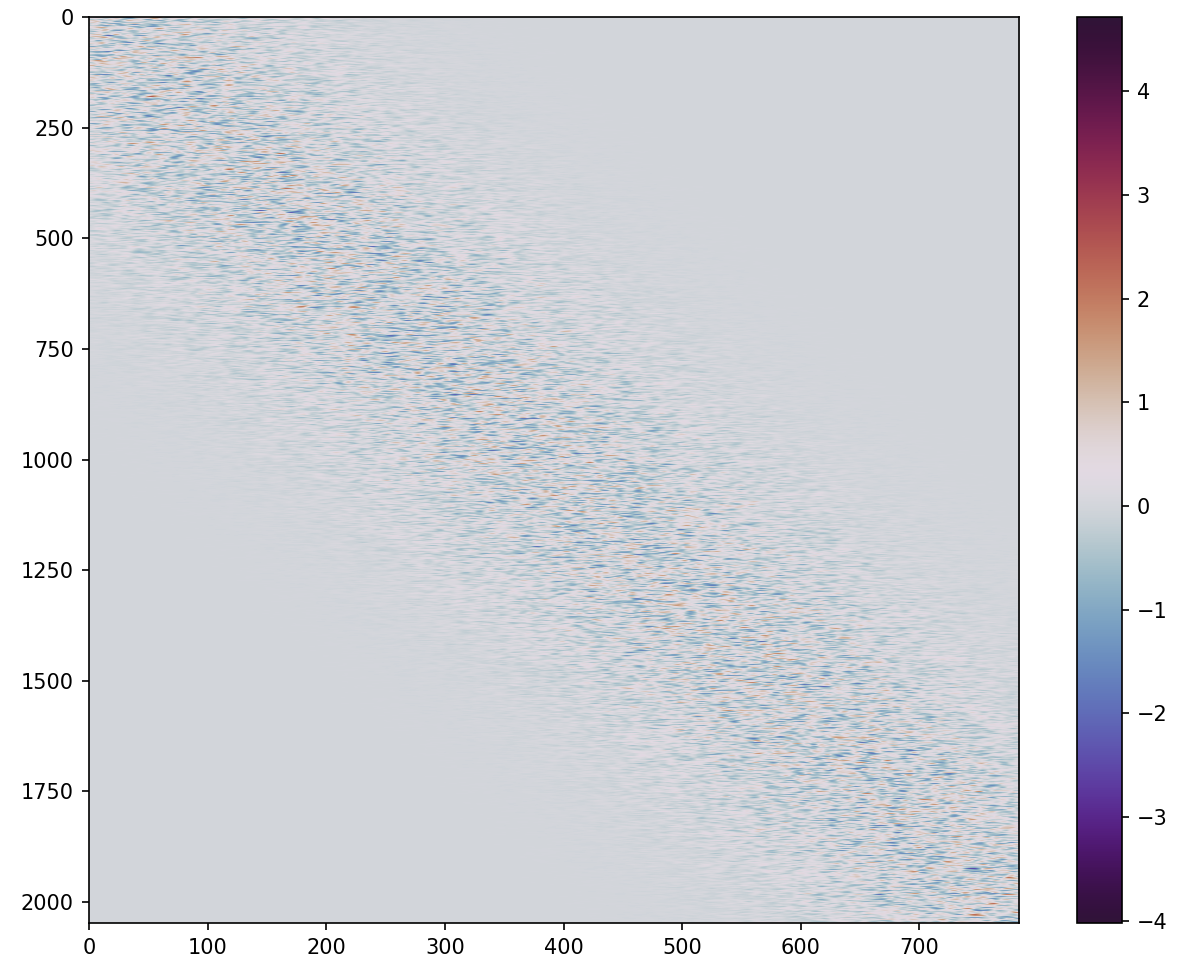}
    \subcaption{Model 1 with \\ $\sigma_r=0.1$, $\sigma_s=0.01$}
  \end{minipage}
  \begin{minipage}[b]{0.33\linewidth}
    \centering
    \includegraphics[keepaspectratio, scale=0.18]{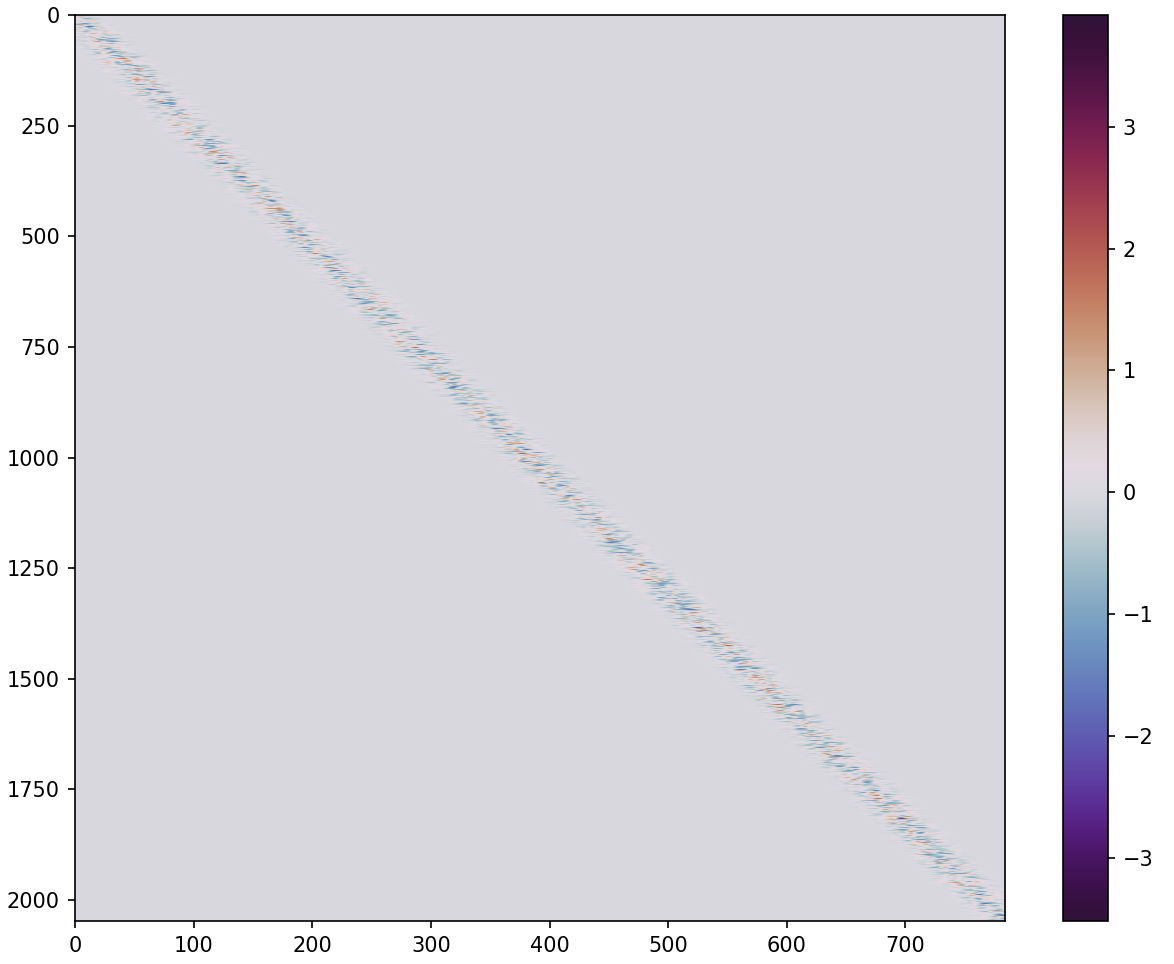}
    \subcaption{Model 1 with \\ $\sigma_r=0.01$, $\sigma_s=0.01$}
  \end{minipage}
  \\
  \begin{minipage}[b]{0.33\linewidth}
    \centering
    \includegraphics[keepaspectratio, scale=0.18]{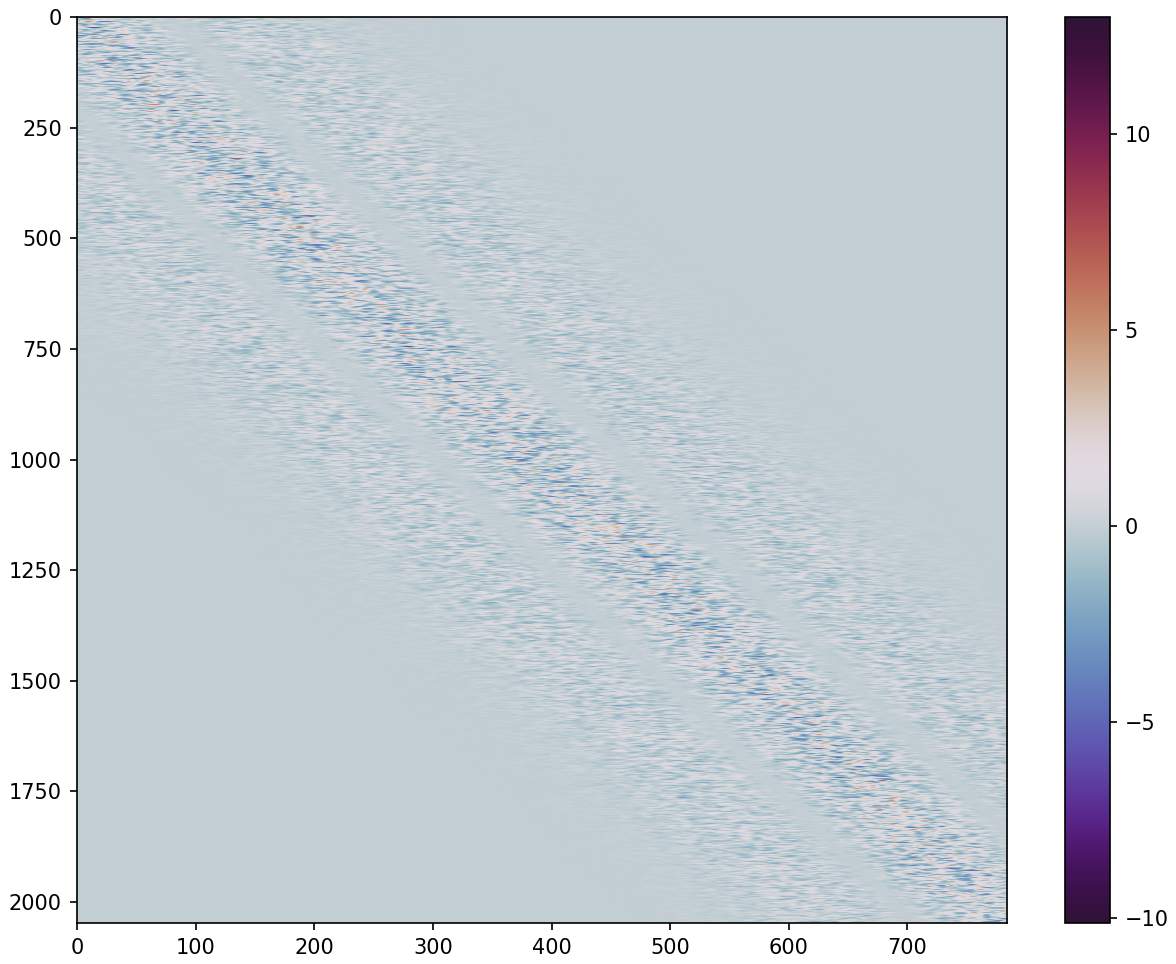}
    \subcaption{Model 3 \\ $\sigma_r=0.1$, $\sigma_s=0.01$}
  \end{minipage}
  \begin{minipage}[b]{0.33\linewidth}
    \centering
    \includegraphics[keepaspectratio, scale=0.18]{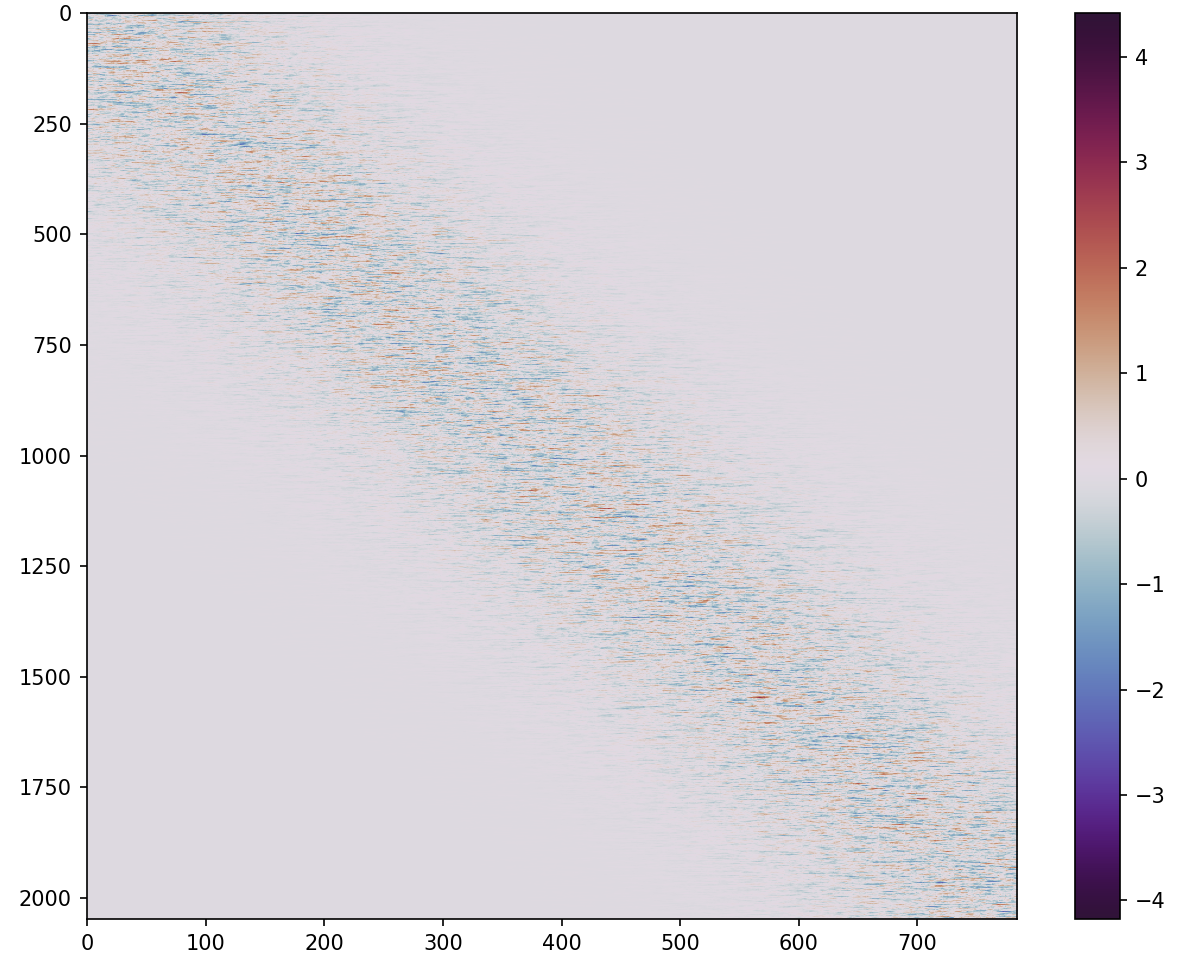}
    \subcaption{Model 4 with \\ $\sigma_r=0.1$, $\sigma_s=0.01$}
  \end{minipage}
  \begin{minipage}[b]{0.33\linewidth}
    \centering
    \includegraphics[keepaspectratio, scale=0.18]{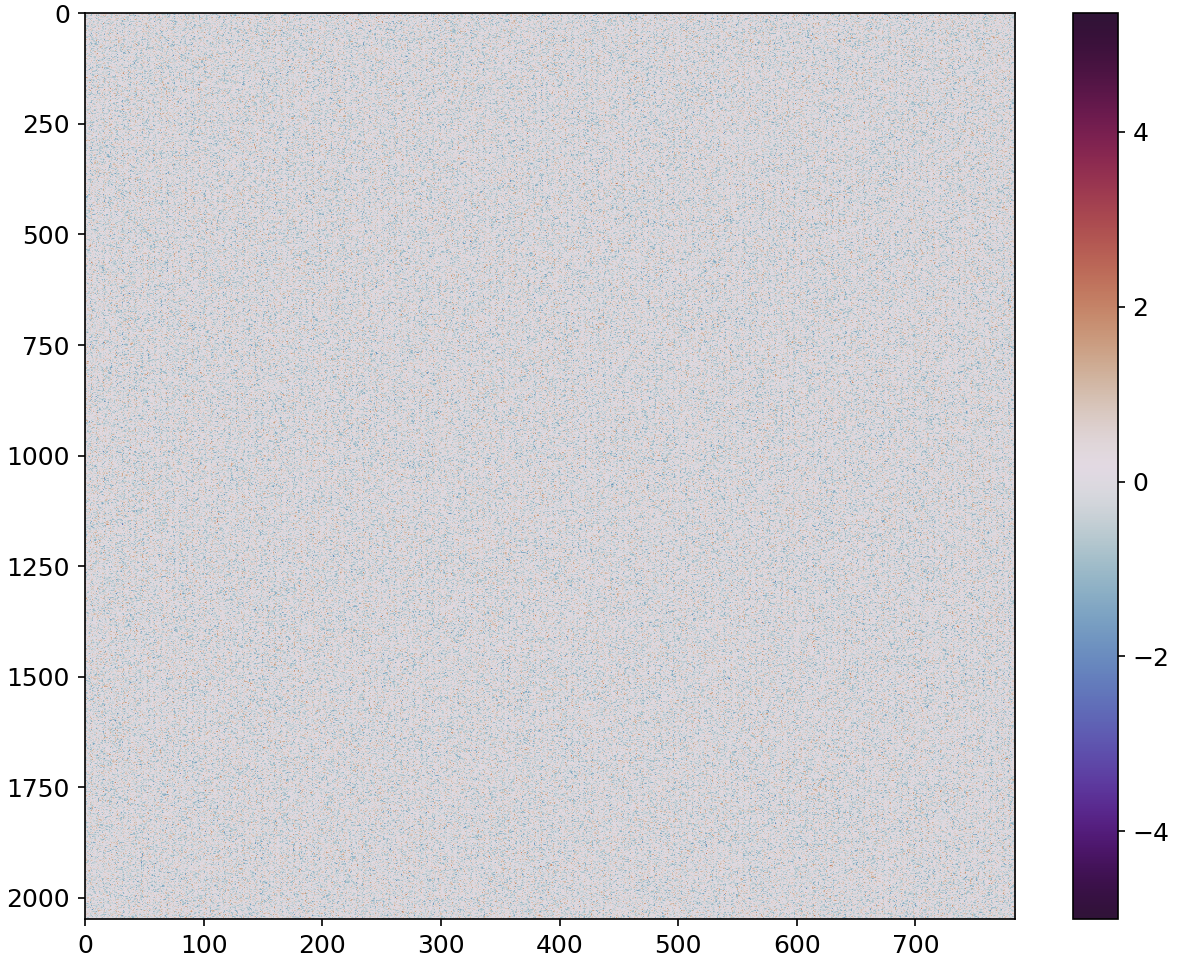}
    \subcaption{Model 5 \\ ~}
  \end{minipage}
  \caption{Initial weight matrix from input layer to first hidden layer. Color bar on right indicates weight values.}
  \label{Initial_weights}
\end{figure}

We visualized the initial weight matrix from the input layer to the first hidden layer of each model defined in Tab. \ref{tableofmodels}. Figs. \ref{Initial_weights}.(a), (b), and (c) show how the receptive field shrinks as $\sigma_r$ is decreased. In Fig. \ref{Initial_weights}.(d), the receptive field is a Mexican hat, and the figure shows the inhibitory property as defined. In Fig. \ref{Initial_weights}.(e), the Mate\'rn kernel is used to correlate between neurons when generating weights, resulting in a difference in distribution. Fig. \ref{Initial_weights}.(f) shows the NTK parameterization, where the weights are generated from a standard Gaussian distribution.

\subsection{Visualization of first-layer output of pre-activation}
\label{fl_output_vis}
\begin{figure}[t]
    \begin{minipage}[b]{0.5\linewidth}
        \centering
        \includegraphics[keepaspectratio, scale=0.6]{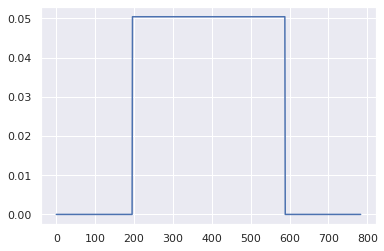}
        \subcaption{Toy data 1.}
    \end{minipage}
    \begin{minipage}[b]{0.5\linewidth}
        \centering
        \includegraphics[keepaspectratio, scale=0.6]{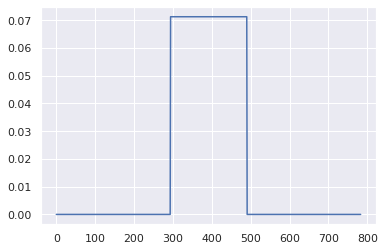}
        \subcaption{Toy data 2.}
    \end{minipage}
    \caption{Input data in the shape of a uniform distribution with dimension 784. Verify for two patterns of different widths to see how the output changes.}
    \label{toydata}
\end{figure}

\begin{figure}[t]
    \begin{minipage}[b]{0.5\linewidth}
        \centering
        \includegraphics[keepaspectratio, scale=0.27]{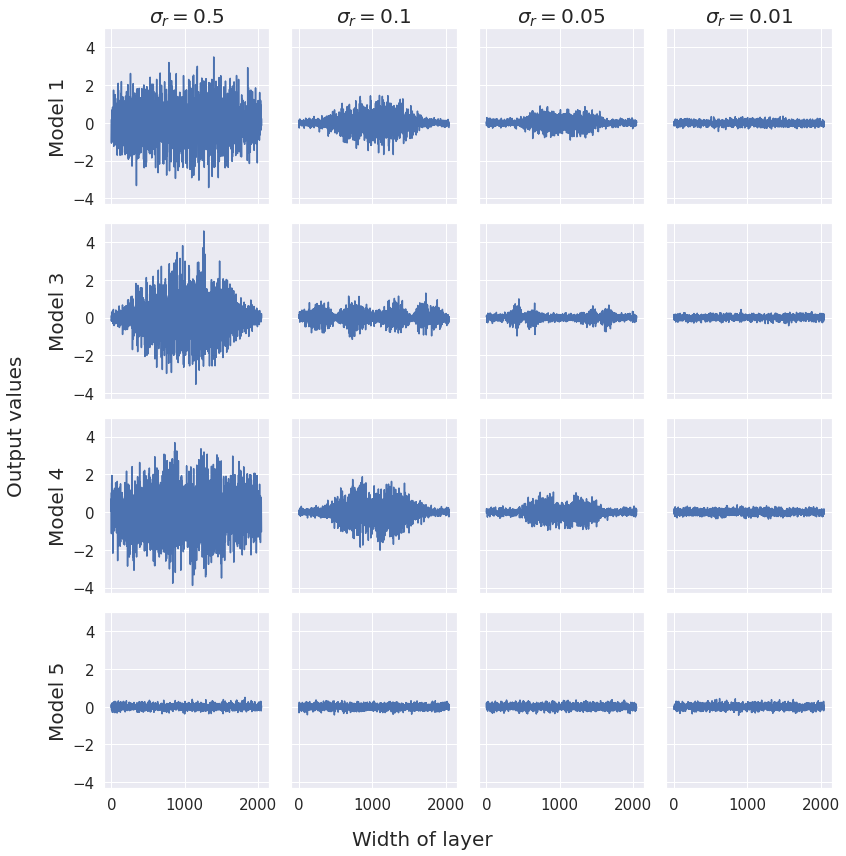}
        \subcaption{Output diagrams in case of toy data 1.}
    \end{minipage}
    \begin{minipage}[b]{0.5\linewidth}
        \centering
        \includegraphics[keepaspectratio, scale=0.27]{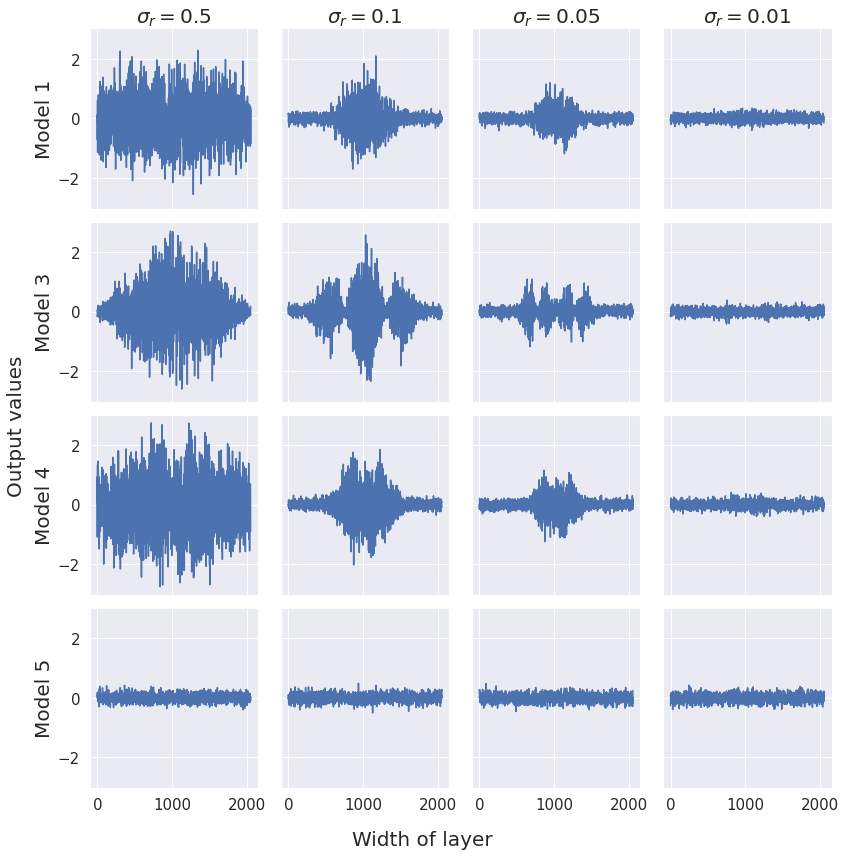}
        \subcaption{Output diagrams in case of toy data 2.}
    \end{minipage}
    \caption{Output values from the first layer when $\sigma_r=0.5, 0.1, 0.05, 0.01$ for Model 1, Model 3, Model 4, and Model 5. In each diagram, the x-axis represents the number of units in the layer, and the y-axis represents the output value before input to the activation function. Model 5 is not affected by $\sigma_r$, however, it is included for comparison.}
    \label{output_diagram}
\end{figure}

The feedforward inhibition of Model 3, which has a Mexican Hat-type receptive field, is difficult to understand just by visualizing the weight matrix. Therefore, we plot the output of pre-activation from the first layer of each model (Model 1, Model 3, Model 4, and Model 5) by varying the variance value $\sigma_r$, which describes the size of the receptive field. Two toy data (Fig. \ref{toydata}) are used as input. Model 3 shows stronger localization than the other models (Fig. \ref{output_diagram}). For example, when $\sigma_r=0.1$, its output is small for toy data 1 (wide data) but large for toy data 2 (localized data).

\subsection{Other models in the NTK regime}
\label{other_ntk_regime}
\begin{figure}[t]
\captionsetup[subfigure]{justification=centering}
  \begin{minipage}[b]{\linewidth}
    \centering
    \includegraphics[keepaspectratio, scale=0.4]{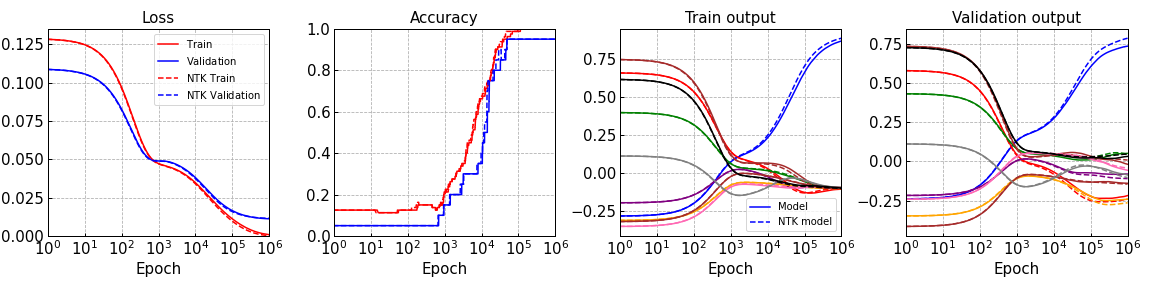}
    \subcaption{Comparison of SGD/NTK-based loss/accuracy/output curves for Model 2 with $\sigma_r=0.5$ and $\sigma_s=0.01$.}
    \label{model2_ntkregime}
  \end{minipage}
  \begin{minipage}[b]{\linewidth}
    \centering
    \includegraphics[keepaspectratio, scale=0.4]{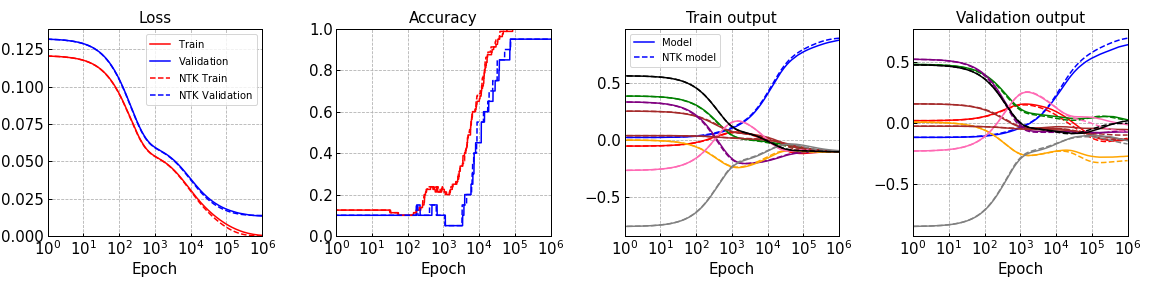}
    \subcaption{Comparison of SGD/NTK-based loss/accuracy/output curves for Model 3 with $\sigma_r=0.01$ and $\sigma_s=0.01$.}
    \label{model3_ntkregime}
  \end{minipage}
  \begin{minipage}[b]{\linewidth}
    \centering
    \includegraphics[keepaspectratio, scale=0.4]{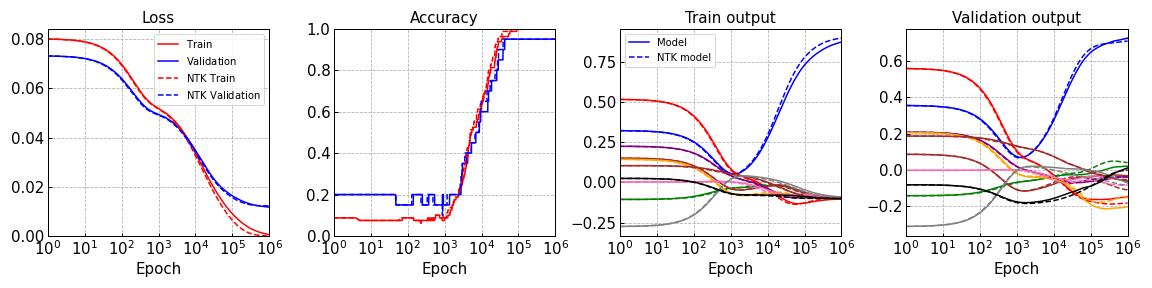}
    \subcaption{Comparison of SGD/NTK-based loss/accuracy/output curves for Model 4 with $\sigma_r=0.5$ and $\sigma_s=0.01$.}
    \label{model4_ntkregime}
  \end{minipage}
  \begin{minipage}[b]{\linewidth}
    \centering
    \includegraphics[keepaspectratio, scale=0.4]{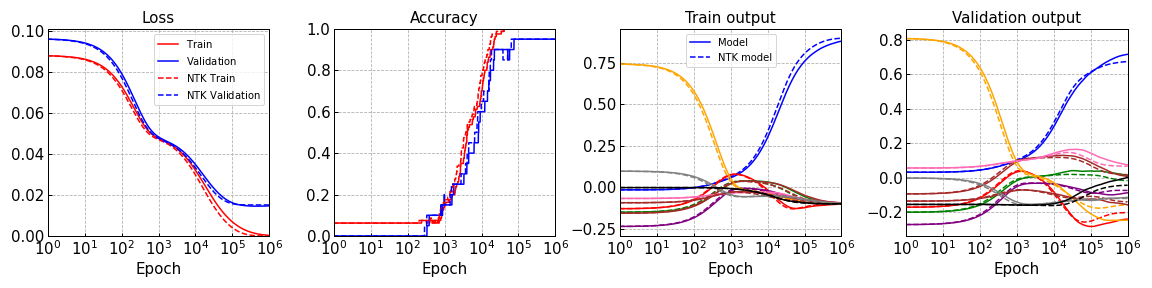}
    \subcaption{Comparison of SGD/NTK-based loss/accuracy/output curves for Model 5.}
    \label{model7_ntkregime}
  \end{minipage}
  \caption{Comparison of SGD/NTK-based loss/accuracy/output curves for five models.}
  \label{all_ntk_regime}
\end{figure}

We confirm that Models 2, 3, 4, and 5 are also in the NTK regime (Fig. \ref{all_ntk_regime}).

\subsection{Calculating loss by NTK regression with different combinations of \texorpdfstring{$\sigma_r$}{sigmar} and \texorpdfstring{$\sigma_s$}{sigmas}}
\label{grids_ntkreg}
\begin{figure}[t]
\captionsetup[subfigure]{justification=centering}
  \begin{minipage}[b]{0.42\linewidth}
    \centering
    \includegraphics[keepaspectratio, scale=0.27]{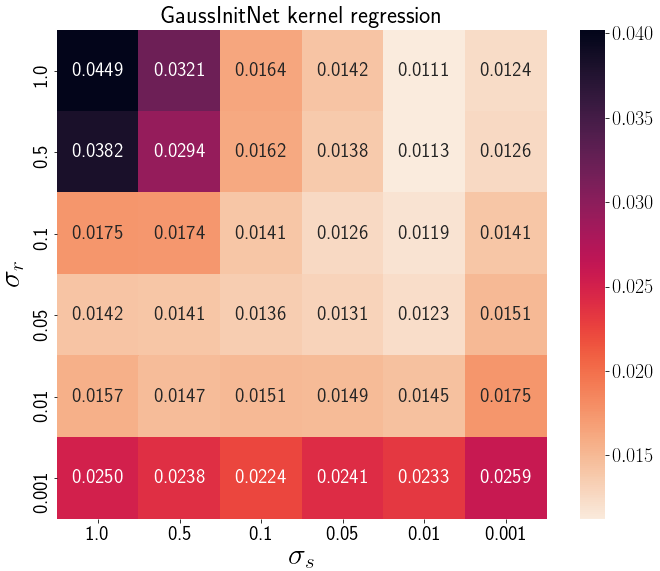}
    \subcaption{Model 1}
  \end{minipage}
  \begin{minipage}[b]{0.6\linewidth}
    \centering
    \includegraphics[keepaspectratio, scale=0.27]{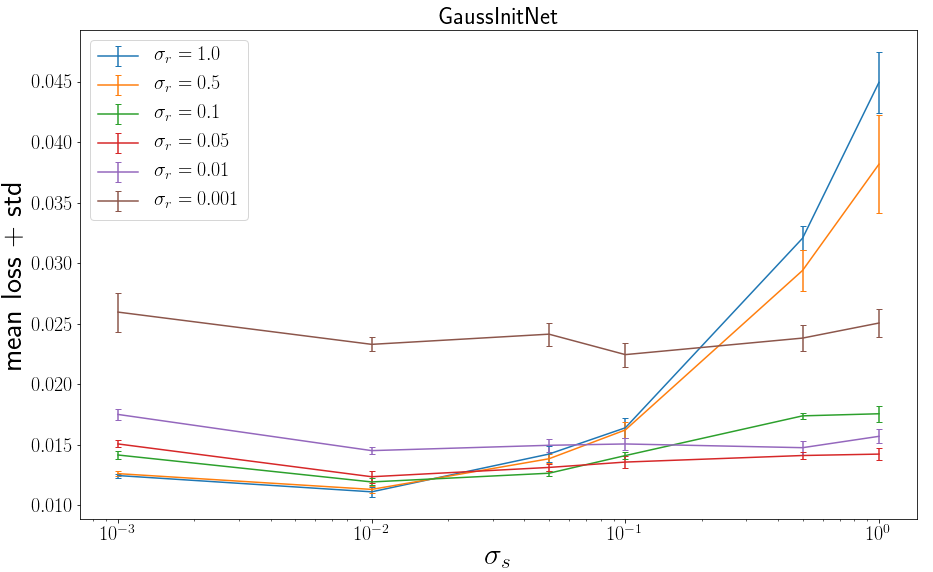}
    \subcaption{Five-time average of loss and its standard deviation}
  \end{minipage}\\
  
  \begin{minipage}[b]{0.42\linewidth}
    \centering
    \includegraphics[keepaspectratio, scale=0.27]{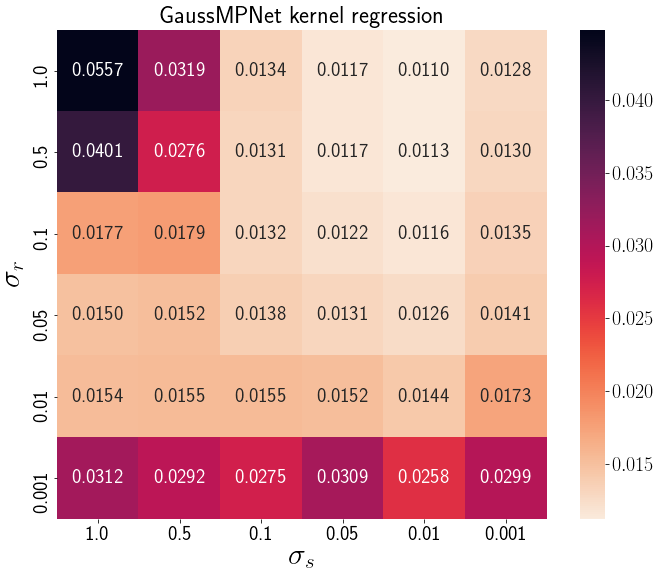}
    \subcaption{Model 2}
  \end{minipage}
  \begin{minipage}[b]{0.6\linewidth}
    \centering
    \includegraphics[keepaspectratio, scale=0.27]{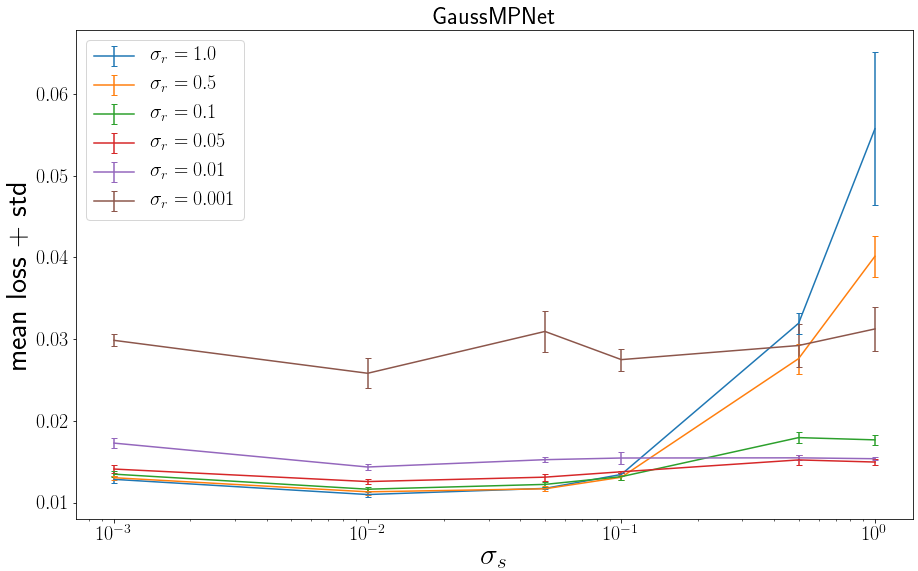}
    \subcaption{Five-time average of loss and its standard deviation}
  \end{minipage}
  \caption{Heatmaps of losses for Model 1 and Model 2, and five-time average and standard deviations of losses with different combinations of $\sigma_r$ and $\sigma_s$.}
  \label{emps_1}
\end{figure}

\begin{figure}[t]
\captionsetup[subfigure]{justification=centering}
  \begin{minipage}[b]{0.42\linewidth}
    \centering
    \includegraphics[keepaspectratio, scale=0.27]{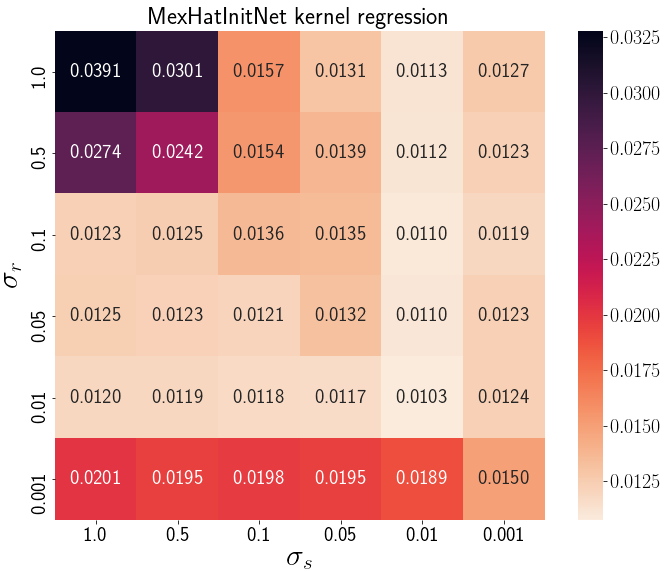}
    \subcaption{Model 3}
  \end{minipage}
  \begin{minipage}[b]{0.6\linewidth}
    \centering
    \includegraphics[keepaspectratio, scale=0.27]{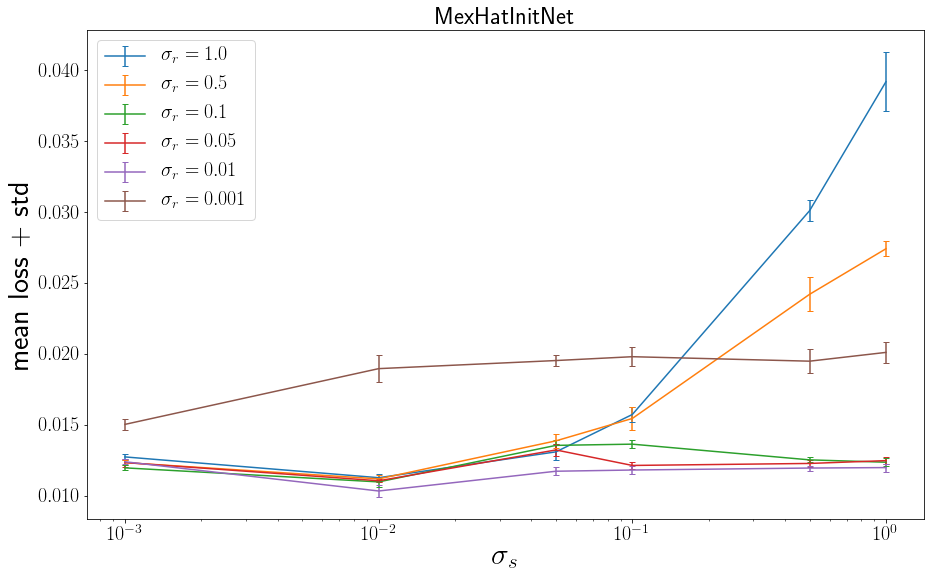}
    \subcaption{Five-time average of loss and its standard deviation}
  \end{minipage}\\
  
  \begin{minipage}[b]{0.42\linewidth}
    \centering
    \includegraphics[keepaspectratio, scale=0.27]{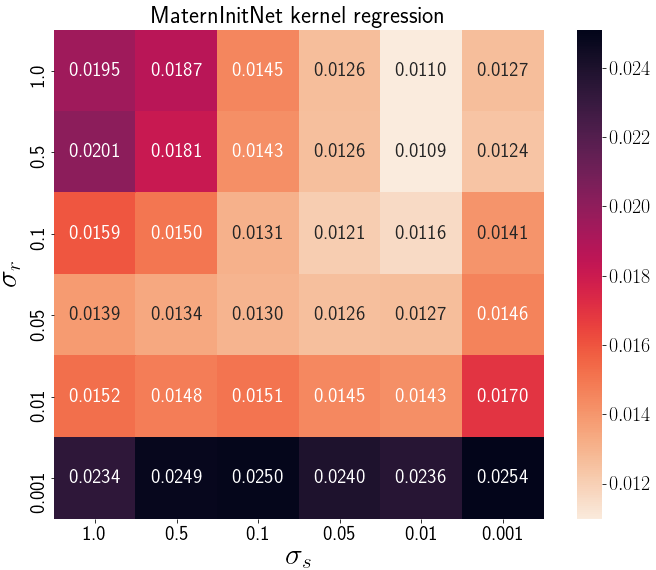}
    \subcaption{Model 4}
  \end{minipage}
  \begin{minipage}[b]{0.6\linewidth}
    \centering
    \includegraphics[keepaspectratio, scale=0.27]{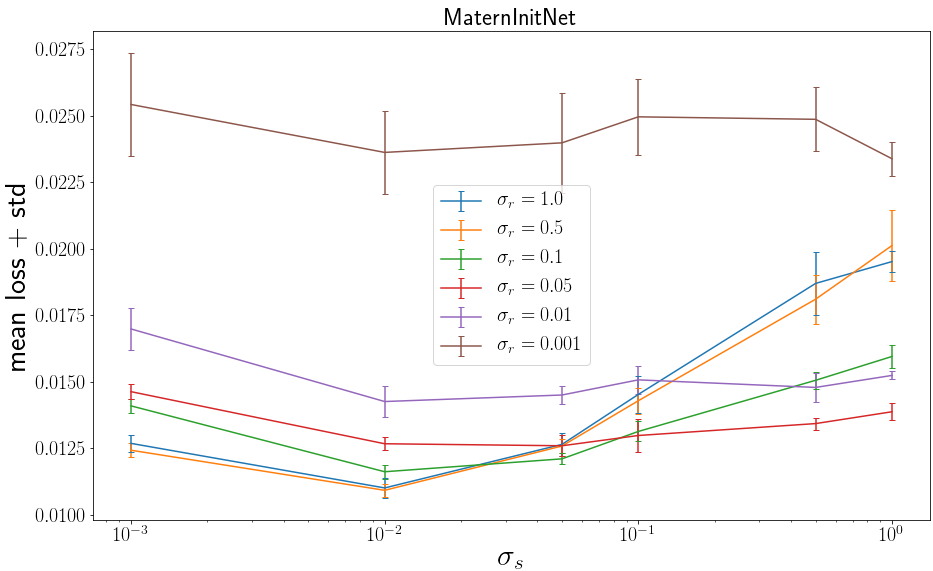}
    \subcaption{Five-time average of loss and its standard deviation}
  \end{minipage}
  \caption{Heatmaps of losses for Model 3 and Model 4, and five-time average and standard deviations of losses with different combinations of $\sigma_r$ and $\sigma_s$.}
  \label{emps_2}
\end{figure}

To verify the validity of the obtained values, the five-time averages of the loss values and their standard deviations were calculated. The results are shown in Figs. \ref{emps_1} and \ref{emps_2}. The left heat map shows the five times average of the loss values corresponding to each $\sigma_r$ and $\sigma_s$. The lighter-colored areas indicate smaller values. The right figure shows the five times average of the loss values and their standard deviations corresponding to each $\sigma_r$ and $\sigma_s$.

\subsection{Average relative distances with different combinations of \texorpdfstring{$\sigma_r$}{sigmar} and \texorpdfstring{$\sigma_s$}{sigmas}}
\label{avg_distance_comb}
\begin{figure}[t]
\captionsetup[subfigure]{justification=centering}
  \begin{minipage}[b]{0.48\linewidth}
    \centering
    \includegraphics[keepaspectratio, scale=0.365]{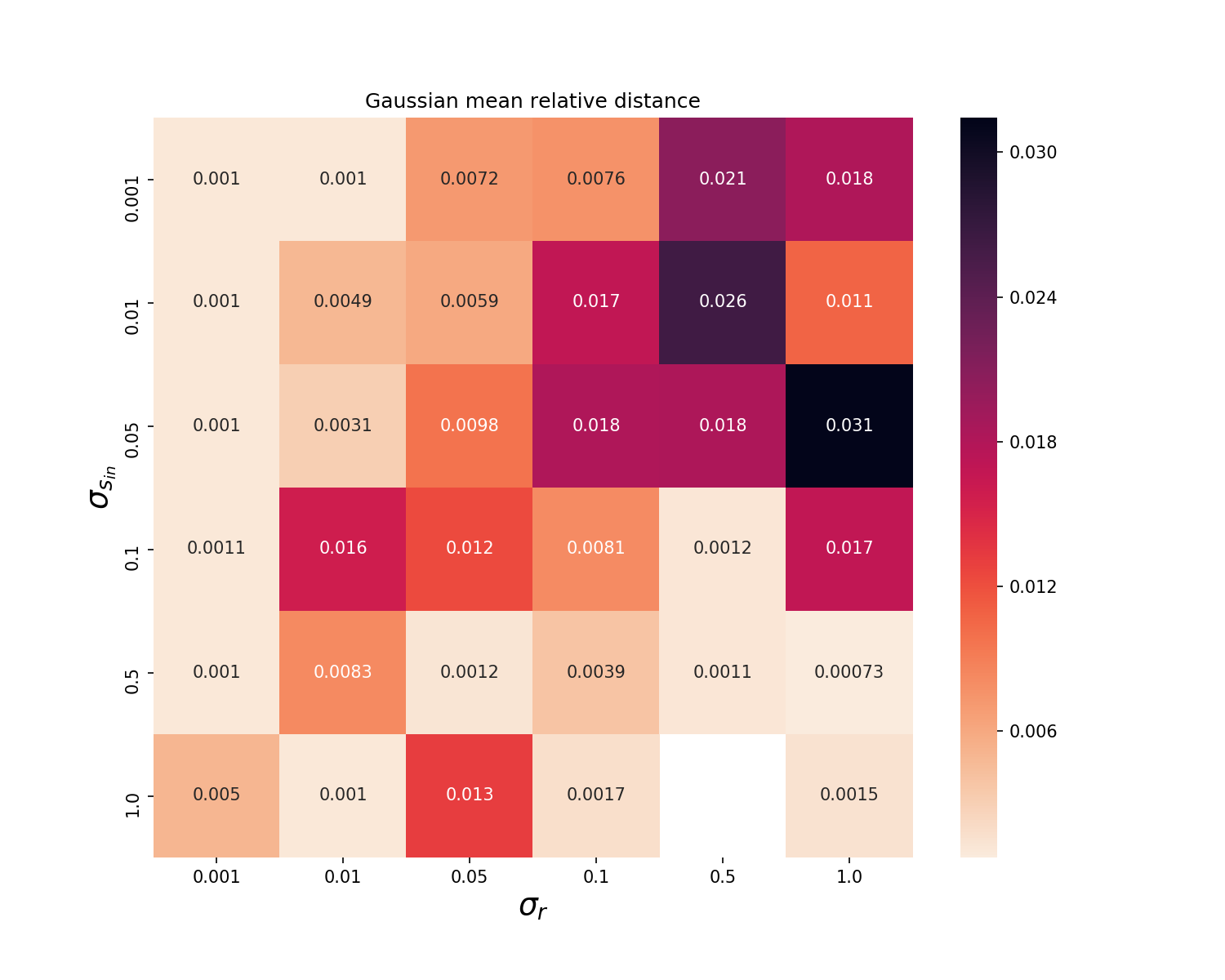}
    \subcaption{Model 1}
  \end{minipage}
  \begin{minipage}[b]{0.5\linewidth}
    \centering
    \includegraphics[keepaspectratio, scale=0.365]{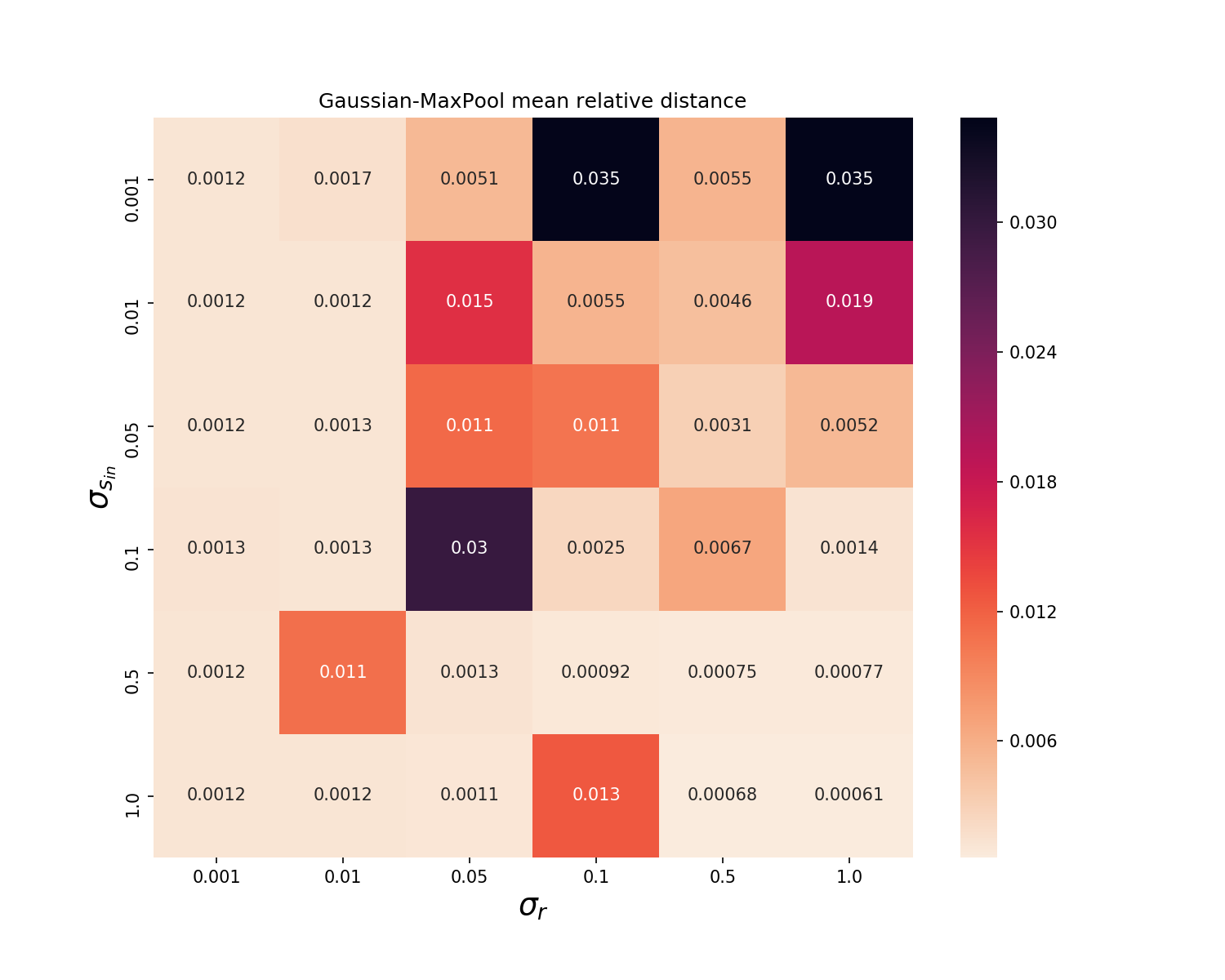}
    \subcaption{Model 2}
  \end{minipage}\\
  
  \begin{minipage}[b]{0.48\linewidth}
    \centering
    \includegraphics[keepaspectratio, scale=0.365]{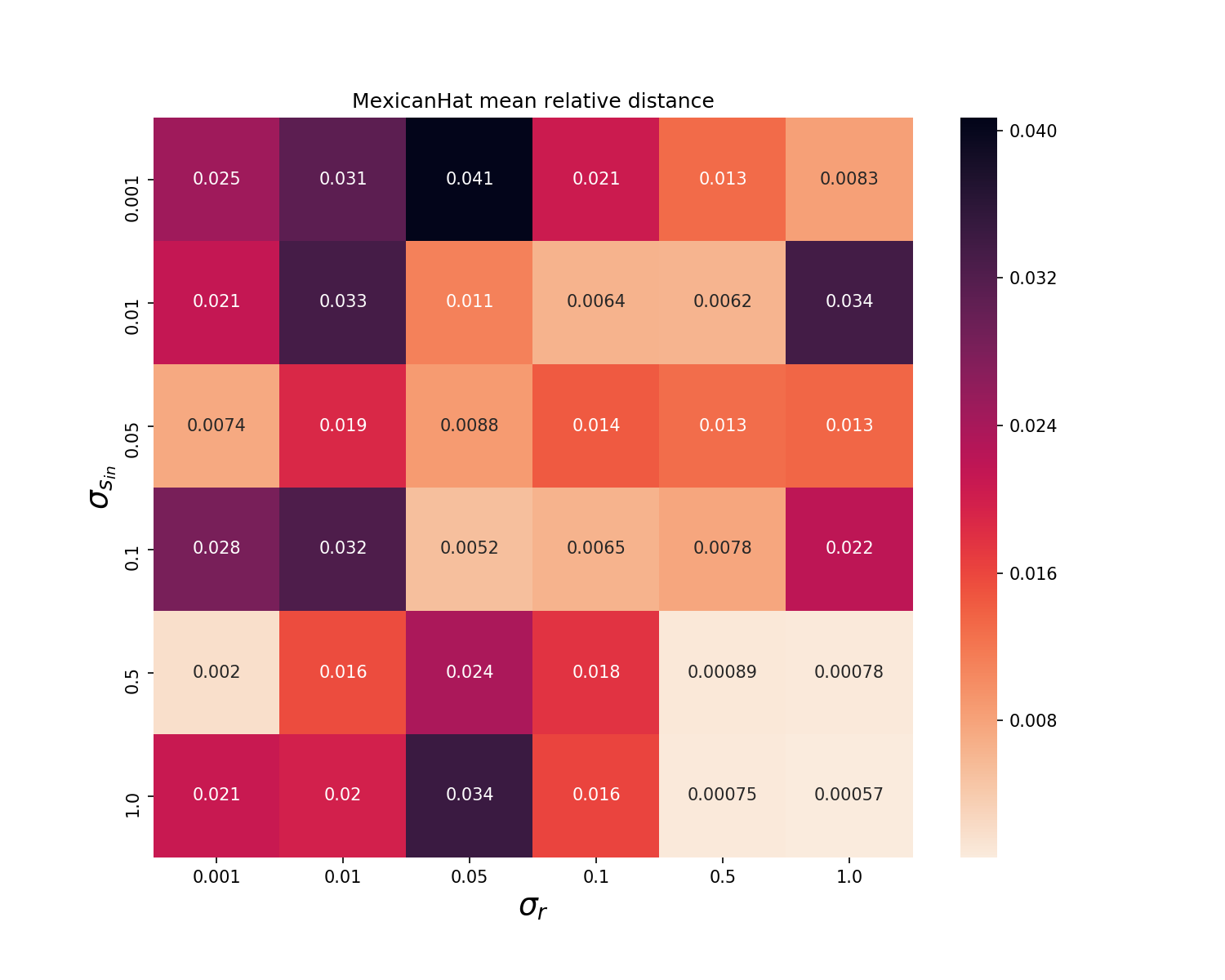}
    \subcaption{Model 3}
  \end{minipage}
  \begin{minipage}[b]{0.5\linewidth}
    \centering
    \includegraphics[keepaspectratio, scale=0.365]{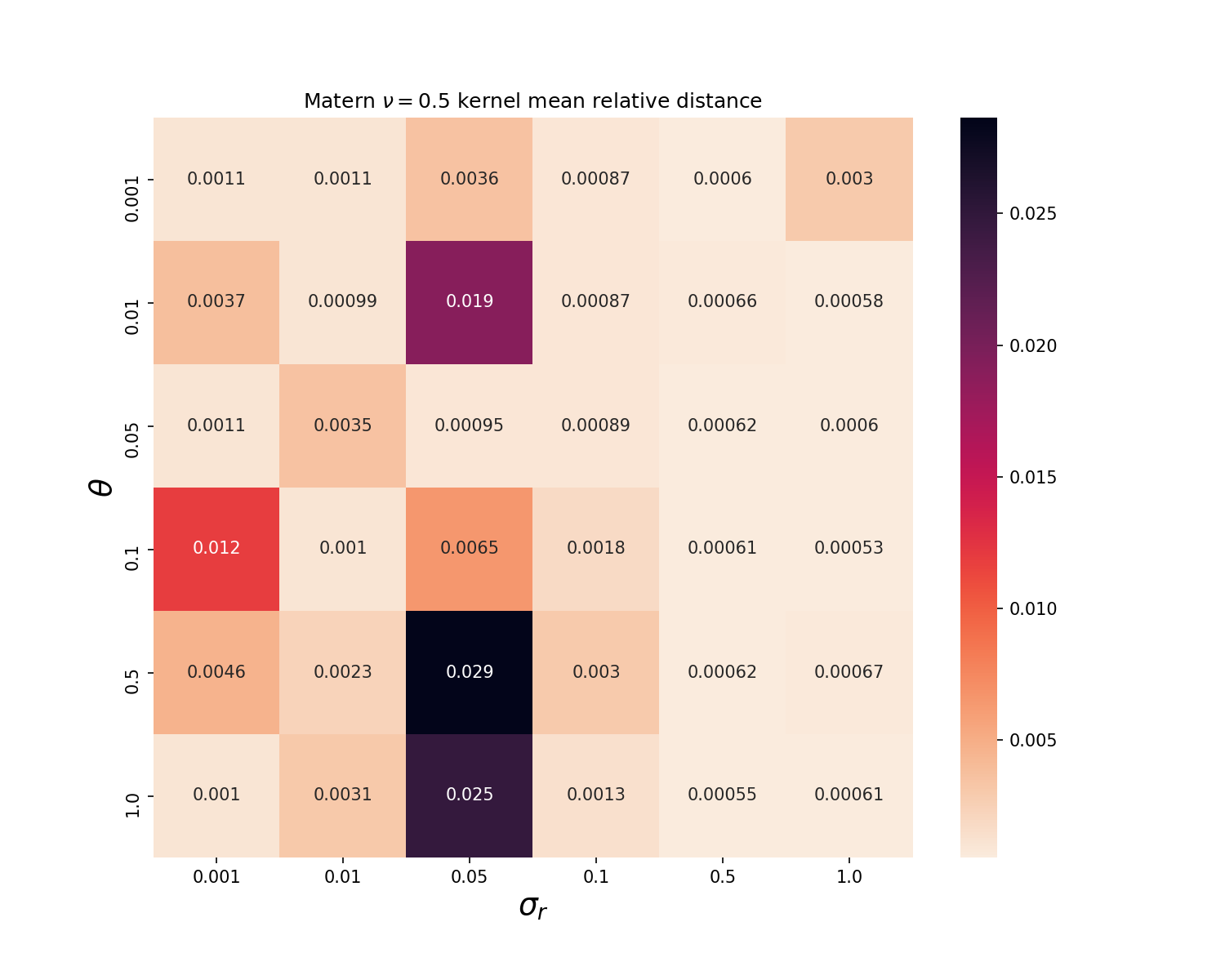}
    \subcaption{Model 4}
  \end{minipage}
  \caption{Average relative distances to deformations with different combinations of $\sigma_r$ and $\sigma_s$.}
  \label{rdtod}
\end{figure}

\begin{figure}[t]
\captionsetup[subfigure]{justification=centering}
  \begin{minipage}[b]{0.48\linewidth}
    \centering
    \includegraphics[keepaspectratio, scale=0.365]{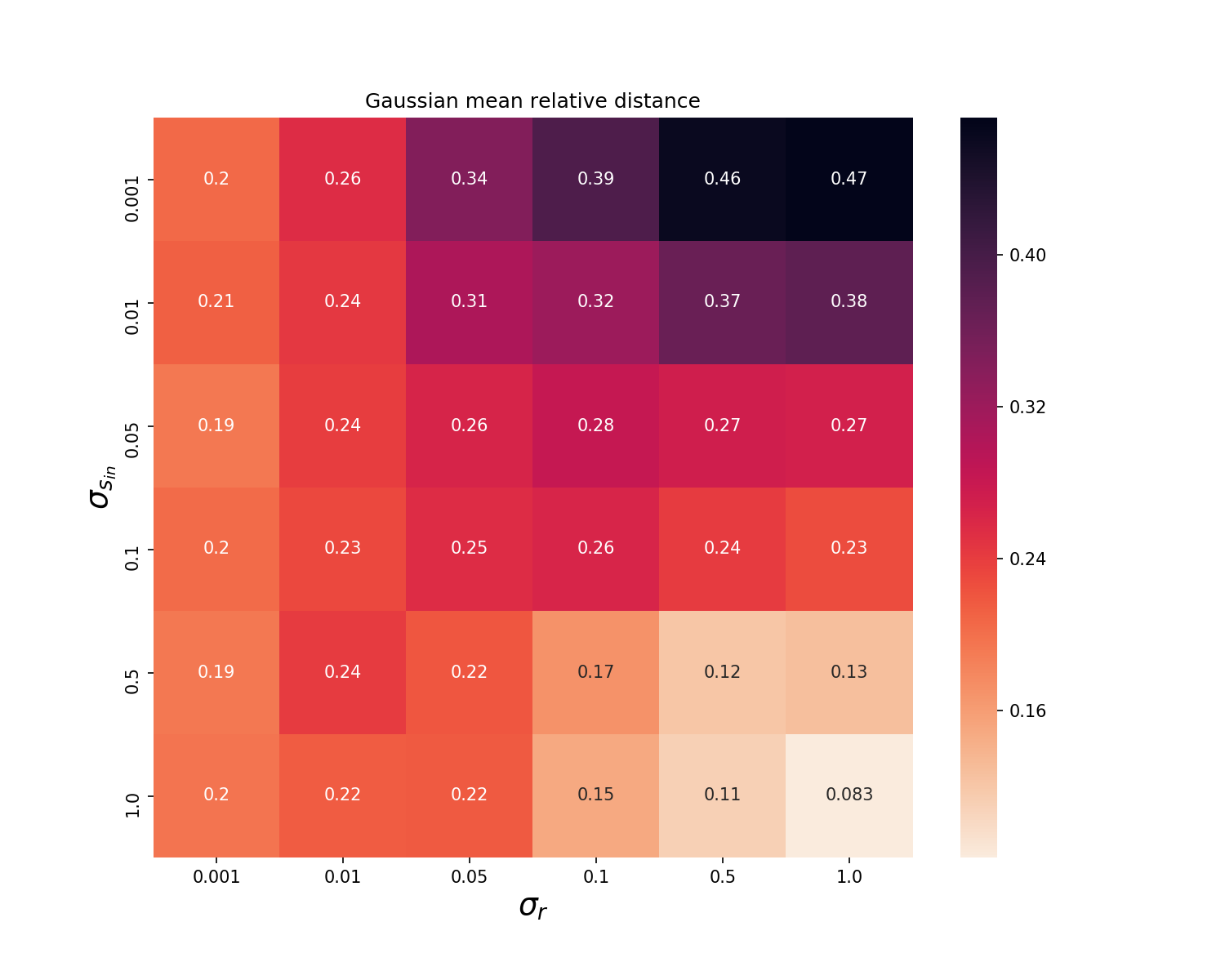}
    \subcaption{Model 1}
  \end{minipage}
  \begin{minipage}[b]{0.5\linewidth}
    \centering
    \includegraphics[keepaspectratio, scale=0.365]{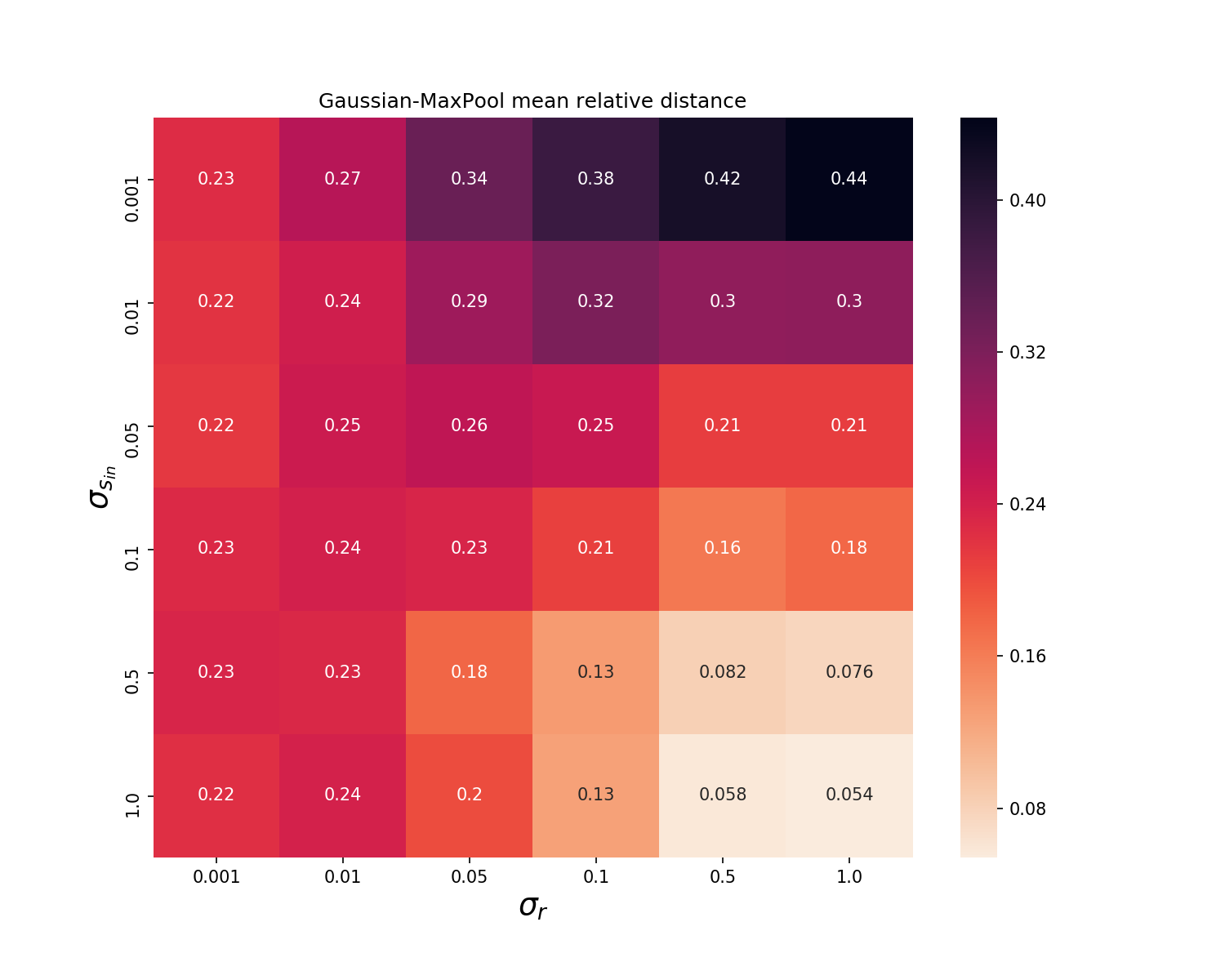}
    \subcaption{Model 2}
  \end{minipage}\\
  
  \begin{minipage}[b]{0.48\linewidth}
    \centering
    \includegraphics[keepaspectratio, scale=0.365]{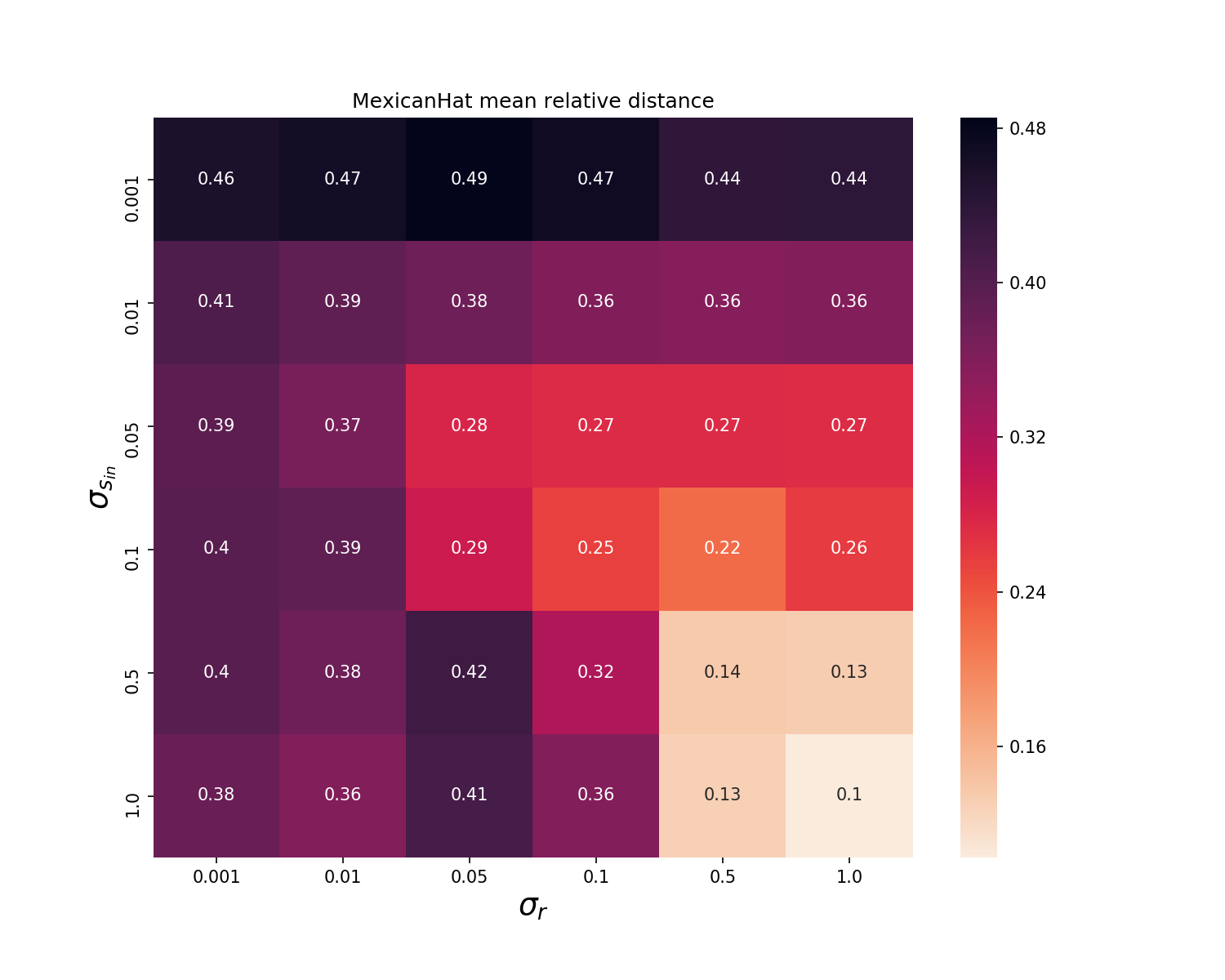}
    \subcaption{Model 3}
  \end{minipage}
  \begin{minipage}[b]{0.5\linewidth}
    \centering
    \includegraphics[keepaspectratio, scale=0.365]{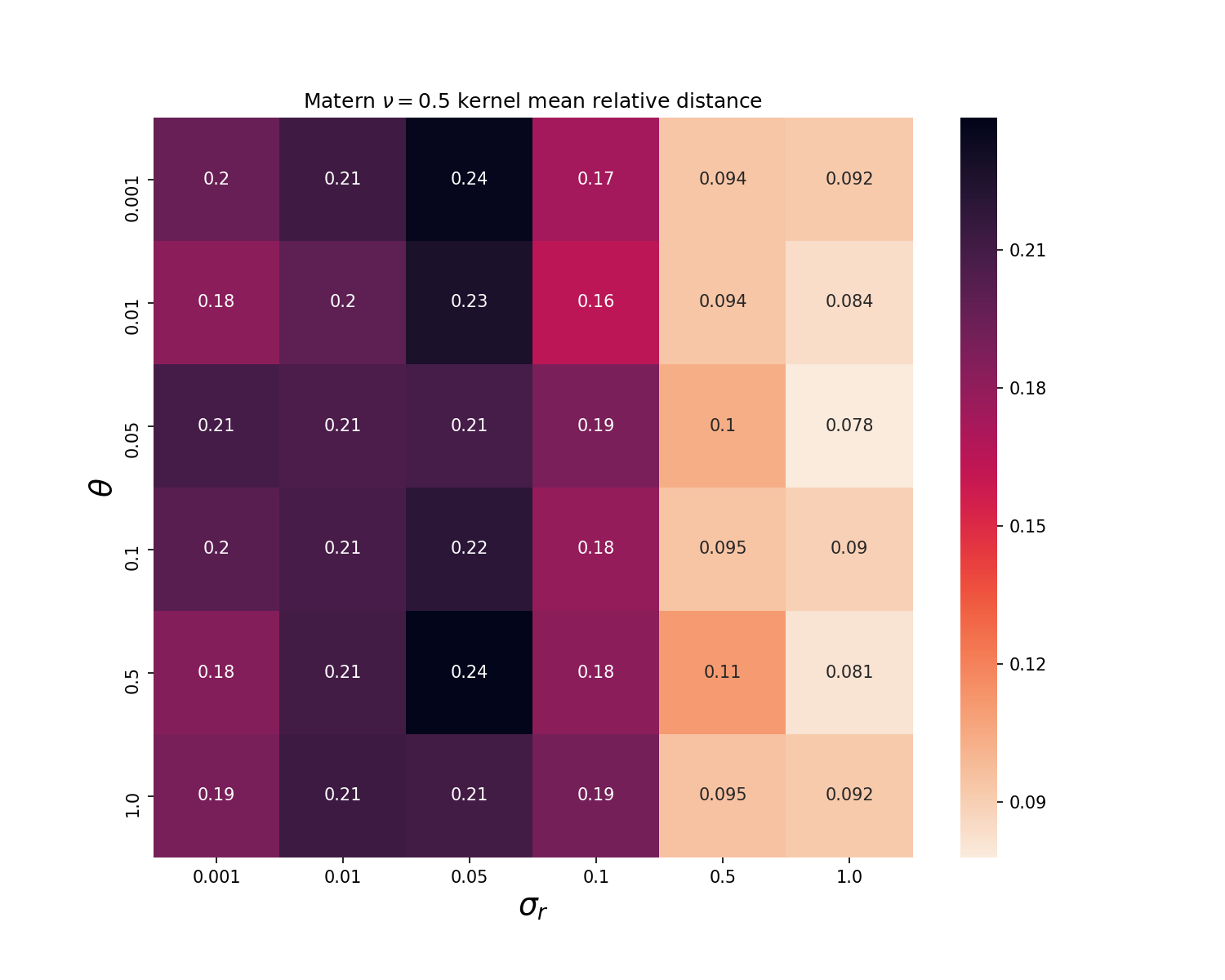}
    \subcaption{Model 4}
  \end{minipage}
  \caption{Average relative distances to translations and deformations with different combinations of $\sigma_r$ and $\sigma_s$.}
  \label{rdtotad}
\end{figure}

To obtain the optimal parameters $\sigma_r$ and $\sigma_s$ that minimize average relative distances to translations and deformations, a parameter search is performed. The results are shown in Figs.\ref{rdtod} and \ref{rdtotad}. These figures shows the average relative distances to translations and deformations corresponding to each $\sigma_r$ and $\sigma_s$. The lighter-colored areas indicate smaller values.

\end{document}